\documentclass[11pt]{article}

\usepackage[utf8]{inputenc}
\usepackage[T1]{fontenc}
\usepackage{lmodern}
\usepackage[margin=1in]{geometry}
\usepackage{setspace}
\usepackage{amsmath,amssymb,amsfonts,amsthm,mathtools}
\usepackage{booktabs}
\usepackage{nicefrac}
\usepackage{microtype}
\usepackage{bbm}
\usepackage{physics}
\usepackage{bm}
\usepackage[short]{optidef}
\usepackage{placeins}
\usepackage{xfrac}
\usepackage{acronym}
\usepackage{graphicx}
\usepackage{xcolor}
\usepackage{multirow}
\usepackage{subcaption}
\usepackage{tikz}
\usetikzlibrary{calc,patterns,arrows,shapes.arrows,intersections}

\usepackage{algorithmic}
\usepackage[ruled,vlined]{algorithm2e}
\SetKwInput{KwInput}{Input}
\SetKwInput{KwOutput}{Output}

\usepackage{natbib}
\bibpunct[, ]{(}{)}{,}{a}{}{,}
\usepackage[hidelinks]{hyperref}

\newtheorem{theorem}{Theorem}
\newtheorem{lemma}{Lemma}
\newtheorem{proposition}{Proposition}
\newtheorem{corollary}{Corollary}

\theoremstyle{definition}
\newtheorem{remark}{Remark}

\title{Optimal Control of Multiclass Fluid Queueing Networks: A Machine Learning Approach}

\author{%
Dimitris Bertsimas\\
Sloan School of Management, Massachusetts Institute of Technology\\
\texttt{dbertsim@mit.edu}
\and
Cheol Woo Kim\\
Operations Research Center, Massachusetts Institute of Technology\\
\texttt{acwkim@mit.edu}
}

\date{}

\begin{document}
\maketitle

\begin{abstract}
We propose a machine learning approach to the optimal control of multiclass fluid queueing networks (MFQNETs) that provides explicit and insightful control policies. We prove that a piecewise constant optimal policy exists for MFQNET control problems, with segments separated by hyperplanes passing through the origin. We use Optimal Classification Trees with hyperplane splits (OCT-H) to learn an optimal control policy for MFQNETs. We use numerical solutions of MFQNET control problems as a training set and apply OCT-H to learn explicit control policies.  Furthermore, we show that both the theoretical results and the proposed algorithm extend to robust MFQNETs with uncertain service and arrival rates. We report experimental results with up to 33 servers and 99 classes that demonstrate that the learned policies achieve 100\% accuracy on the test set.  While the offline training of OCT-H can take days in large networks, the online application takes milliseconds.
\end{abstract}


%%%%%%%%%%%%%%%%%%%%%%%%%%%%%%%%%%%%%%%%%%%%%%%%%%%%%%%%%%%%%%%%%%%%%%

% Samples of sectioning (and labeling) in OPRE
% NOTE: (1) \section and \subsection do NOT end with a period
%       (2) \subsubsection and lower need end punctuation
%       (3) capitalization is as shown (title style).
%
%\section{Introduction.}\label{intro} %%1.
%\subsection{Duality and the Classical EOQ Problem.}\label{class-EOQ} %% 1.1.
%\subsection{Outline.}\label{outline1} %% 1.2.
%\subsubsection{Cyclic Schedules for the General Deterministic SMDP.}
%  \label{cyclic-schedules} %% 1.2.1
%\section{Problem Description.}\label{problemdescription} %% 2.

% Text of your paper here

\section{Introduction}
\label{sec:intro}
Multiclass queueing networks are complex systems that model the behavior of multiple classes of jobs, each with their own arrival and service rates, routing paths and holding costs. These networks find numerous applications in diverse fields, including manufacturing \citep{kumar_re-entrant_1993-1}, healthcare \citep{healthcare} and communication networks \citep{comminucation} among many. The control of multiclass queueing networks is of great importance in improving system efficiency, optimizing resource allocation, and reducing operational costs. However, the inherent complexity of these systems makes their analysis and control a challenging task. 

Multiclass fluid queueing networks (MFQNETs) have been developed as a deterministic, continuous approximation of multiclass queueing networks, primarily to provide a tractable method for analyzing the stability of the underlying multiclass queueing networks.  \citep{dai1995, Stolyar1995} demonstrate that the stability of MFQNETs implies the stability of underlying multiclass queueing networks. Several related studies including \citep{Meyn1995, dumas1999, gamarnikhans2005} have also explored the topic.

MFQNETs also provide a useful way to construct control policies for the underlying stochastic multiclass queueing networks. Many approaches have been proposed in the literature to translate the optimal policy for MFQNETs into an effective control policy for their stochastic counterparts \citep{ meyn1997stability, Mag1999, Bauerle2000, Mag2000, robustfluid}. These methods demonstrate strong empirical performance, often supported with theoretical guarantees. In many cases, the optimal policy for MFQNETs also provides structural insights into the optimal policy for their stochastic counterparts \citep{avrambertsimasricard, meyn03, piu09}. However, although MFQNETs are developed as approximations of stochastic networks, solving MFQNET control problems is still computationally intensive. As a result, these methods are generally limited to small instances in practice. For a comprehensive review of the topic, see \citep{Meyn2007} and  \citep{queuebook}.

%\citep{Mag1999, Mag2000} propose discrete review policies and show that they achieve asymptotic optimality and stability under fluid scailing. \citep{robustfluid} provide a robust formulation of MFQNET control problem and translate the resulting policy to the underlying multiclass queueing network. \citep{BertSethu2002, DaiWeiss2002} propose methods to approximately minimize make-span based on the associated fluid models.

Mathematically, optimal control of MFQNETs falls into a subclass of infinite dimensional linear optimization models known as separated continuous linear programs (SCLPs). Several researchers have investigated the theoretical properties of SCLPs, such as \citep{AndNashPer1983, Pullan1995, Pullan1996, Pullan1997}.
 \citep{avrambertsimasricard} find closed-form optimal policies for specific MFQNETs using optimality conditions from optimal control theory. Other works have proposed numerical algorithms for solving SCLPs. \citep{Pullan1993a, Pullan2002, LuoBertsimas} develop algorithms based on discretization, while \citep{Weiss2008ASB, evgenyweiss2021} propose simplex-like methods.  \citep{FleiSethu2005, BampouKuhn2012} propose polynomial-time approximation algorithms.

Despite significant efforts in the field and its practical applications, the optimal control of MFQNETs remains a challenging computational task. Moreover, in practice, one often needs to solve MFQNET control problems with different initial states repeatedly. Since numerical algorithms for SCLPs generate only a single trajectory from a specific initial state, addressing multiple instances requires repeated use of these algorithms. This approach is intractable for large networks and are not feasible in real-time settings. Therefore, computing an optimal policy that maps the state vector to the corresponding optimal control vector across the entire state space is possibly the only computationally tractable alternative.

Recently, there has been growing interest in applying machine learning techniques to solve challenging optimization and control problems. For instance, \citep{Khalili2016, Alvarez2017AML,OPT, Voice, Prune, cauligi2021coco} propose machine learning-based approaches to mixed-integer optimization. \cite{ARO} develop a method to solve two-stage adaptive robust optimization problems using machine learning. Machine learning has been used for hyperparameter tuning in optimization algorithms as well \citep{Hutter2011, Balcan2020LearningTO}. For queueing network control, reinforcement learning methods are proposed in \citep{Raeis2021QueueLearningAR, Bai2019, Dai2022}. See also \citep{bps18, dk18}, where interpretable reinforcement learning methods are proposed. Although these approaches have shown to be effective in addressing computational challenges, it is often difficult to provide theoretical guarantees that the machine learning methods {\color{black}can learn exact optimal policies.}

In this paper, we present a novel approach that leverages machine learning to solve MFQNET control problems  with varying initial states. The MFQNET control problem we consider is the fluid analog of the sequencing problem in multiclass queueing networks. The sequencing problem in multiclass queueing networks is a stochastic and discrete control problem that involves deciding which class of jobs to process at each server at any given time, with the aim of minimizing the expected total cost. We formulate the fluid analog of this problem as a SCLP problem and propose a machine learning-based algorithm to address it.

We solve multiple MFQNET control problems and use the resulting numerical solutions to learn an optimal policy. The machine learning algorithm we use is Optimal Classification Trees with hyperplane splits (OCT-H) proposed by \citep{OCT, MLOPT}. OCT-H is a classification algorithm that partitions the feature space using hyperplanes and assigns a prediction to each region. We prove that OCT-H can learn exact  optimal policies for the MFQNET control problems.

%Given a training set where each data point consists of a covariate vector and an associated target, OCT-H learns a near-optimal decision tree for classification tasks. For each node split, it uses arbitrary linear combination of the features, as opposed to using a single future which is more common form of decision trees. 

\noindent
The contributions of the paper are as follows.
\begin{enumerate}
    \item We prove the existence of a piecewise constant optimal policy for MFQNET control problems, with segments separated by hyperplanes passing through the origin.  This result was previously proven only for special cases.
    \item Based on the theoretical findings, we propose an efficient algorithm that can learn a closed-form optimal policy for MFQNETs using OCT-H. We report 
experimental results with up to 33 servers and 99 classes that demonstrate that the learned policies achieve 100\% accuracy on the test set.  In other words, given any new state, the learned policies are able to consistently output the corresponding optimal control.
    \item Once a policy is learned offline, it can be directly applied  online to unseen states in milliseconds, leading to a significant speed-up compared to solving the problem numerically.
    \item The high interpretability of decision trees allows us to gain insights into the structure of the optimal policy, which is another significant advantage that numerical optimization algorithms often lack. By providing the actual decision trees learned by OCT-H, we develop a deeper understanding of MFQNETs and their optimal policy.
    \item We show that the theoretical results and the proposed algorithm also apply to robust MFQNETs with uncertain service and arrival rates. Our experiments demonstrate that the policies derived from the robust MFQNETs are more robust to parameter perturbations compared to those for the deterministic MFQNETs. The trade-off is that the robust policies are more conservative, which leads to suboptimal performance when the level of uncertainty is low. Additionally, the training data generation requires slightly more computational effort.
\end{enumerate}

To the best of our knowledge, this is the first instance in the reinforcement learning literature where a machine learning algorithm is proven to learn the exact optimal policy, not merely an approximation. This bridges a crucial gap between machine learning methods and exact optimal control in complex systems.

Furthermore, our approach provides a tractable method to solve the otherwise intractable MFQNET control problem exactly in real time. By computing the optimal policy offline — which may be computationally intensive but is still feasible within a reasonable time frame — we can then apply this policy online to determine optimal control actions extremely efficiently. Achieving exact optimal control for MFQNETs in real-time is, to our knowledge, novel and represents a significant advancement in both the theory and practice of queueing network control.

The structure of this paper is as follows. Section \ref{sec:background} provides the definition of MFQNET control, along with the associated optimality conditions. We also provide a brief review of OCT-H  and robust MFQNETs. Section \ref{sec:main} provides our theoretical results on the structure of optimal policy for MFQNET control problems. We develop a learning algorithm based on OCT-H and provide a small example to illustrate the method. Then, we extend the results to robust MFQNETs. Section \ref{sec:numerical experiment} reports the results of computational experiments. In Section \ref{sec:conclusion}, we provide our conclusions.

 \paragraph{Notational conventions} Throughout this paper, we use lower case boldface letters to denote vectors and upper case boldface letters to denote matrices. The $i_{th} $ entry of a vector $\bm{x}$ is denoted $x_i$ or $[\bm{x}]_i$, and the entry in the $i_{th}$ row and $j_{th}$ column of a matrix $\bm{A}$ is denoted $a_{ij}$. Division between two vectors is always assumed to be entry-wise. We use $\bm{e}$ to denote the vector of all ones and $\bm{0}$ to denote the vector of zeros. We use $x(\cdot)$ to denote a real-valued function, and $\bm{x}(\cdot)$ to denote a vector whose entries are real-valued functions. We use $\bm{x}$ instead of $\bm{x}(\cdot)$ when it is clear from the context that $\bm{x}$ is referring to a vector of functions.  We use the expression $\dot{x}(t) \triangleq \frac{dx(t)}{dt}$, the derivative of  $x(t)$ with respect to $t$.

\section{Background}
\label{sec:background}

In this section, we first define the optimal control problem for MFQNETs. Then, we review its key theoretical properties and provide a brief overview of OCT-H.

\subsection{Optimal Control of MFQNETs}
\label{sec:fluid}
Consider a fluid queueing network with $m$ servers and $n$ classes. Each class $i \in [n]$ fluids are processed by a single server $s(i) \in [m]$ with service rate $\mu_i$. After fluids of class $i$ are processed, they either leave the system or change to a different class in a deterministic manner. Fluids may arrive from either another server or from outside the system with external arrival rate $\lambda_i$. If there is no external arrival for class $i$, then $\lambda_i = 0$. The cost per unit time for holding a fluid of class $i$ is denoted $c_i$. 

For each class $i \in [n]$, the control variable $u_i(t)$ denotes the fraction of effort the server $s(i)$ spends processing class $i$ fluids at time $t$. The state variable $x_i(t)$ is the amount of fluids of class $i$ at time $t$.  The dynamics of the system can be expressed using a matrix $\bm{A} \in \mathbb{R}^{n \times n}$, where $a_{ii} = - \mu_i$ and ${a}_{ij} = \mu_j$ if class $i$ receives arrivals from class $j, \; j \neq i$. The rest of the entries of $\bm{A}$ are zero. Then, the dynamics of the system is
$$
\dot{\bm{x}}(t) = \bm{A}\bm{u}(t) + \bm{\lambda}.
$$
\noindent In addition, the sum of the control variables for all classes that are processed at the same server should be less than or equal to one. This constraint can be expressed as
$$
\bm{D}\bm{u}(t) \leq \bm{e},
$$
\noindent where $\bm{D} \in \{0,1\}^{m \times n}$ is a binary matrix with $d_{ji} = 1$ if $s(i) = j$ and $d_{ji} = 0$, otherwise. 

The MFQNET control problem aims to find a control $\bm{u}$ that minimizes the total holding cost of the fluids in the system over the time interval $[0,T]$. We define the MFQNET control problem with an initial state $\bm{x}_0$ as the following:
\begin{alignat}{2}
&\underset{\bm{u}(\cdot), \bm{x}(\cdot)}\min\quad && \int_{0}^{T} \bm{c}^{\top}\bm{x}(t) \,dt  \label{eq:fluid}\\
&s.t. && \dot{\bm{x}}(t) = \bm{A}\bm{u}(t) + \bm{\lambda}, \quad \forall t \in [0,T], \nonumber\\
&\qquad \ && \bm{D}\bm{u}(t) \leq \bm{e}, \quad \forall t \in [0,T],  \nonumber \\
&\qquad \ && \bm{u}(t), \bm{x}(t) \geq \bm{0}, \quad \forall t \in [0,T], \nonumber \\
&\qquad \ && \bm{x}(0) = \bm{x}_0. \nonumber
\end{alignat}
We use $\bm{u}^{*}_{\bm{x}_0}$ and $\bm{x}^{*}_{\bm{x}_0}$ to denote the optimal control and the associated state trajectory of problem \eqref{eq:fluid} with the initial state $\bm{x}_0$. We use $\bm{u}^*$ and $\bm{x}^*$ to denote the optimal control and the associated state trajectory of general MFQNET control problems when the initial state is not specified. 

We define the vector load $-\bm{D}\bm{A}^{-1}\bm{\lambda} \in \mathbb{R}^{m}$, and assume that all its entries are strictly smaller than 1 for stability.  We further assume  $T$ is large enough, so that the system can be emptied by time $T$ \citep{Meyn2007}.  Under this setting, a stationary optimal policy exists. This implies that identifying the optimal initial control $\bm{u}^*_{\bm{x}_0}(0)$ for any initial state $\bm{x}_0$ is equivalent to identifying the optimal control $\bm{u}^*(t)$ for any state $\bm{x}(t)$ at any $t \in [0,T]$.

For each fluid class $i \in [n]$, we define the depletion time $T_i = \inf\{t \in (0,T]: x^*_i(t) = 0\}$ under an optimal state trajectory $\bm{x}^*$, assuming that $x_i^*(0) > 0$. We define the value function $V(\bm{x}_0)$ as the optimal objective value of Problem \eqref{eq:fluid} associated with the initial state $\bm{x}_0$.

\subsection{Properties of MFQNET Control problems}
\label{sec:conditions}
We present several theoretical properties of MFQNET control problems, which will be used to derive our main results. The Pontryagin Maximum Principle \citep{Bittner1963LSP, Sethi2019} provides necessary optimality conditions for general optimal control problems. Due to the non-negativity constraints on the state variable in Problem \eqref{eq:fluid}, the conditions that we provide are tailored for the optimal control problems with pure state constraints. 

We define the Hamiltonian of Problem \eqref{eq:fluid} as
$$
H(\bm{x}, \bm{u}, \bm{y}, t) = \bm{c}^{\top}\bm{x}(t) + \bm{y}(t)^{\top}[\bm{A}\bm{u}(t) + \bm{\lambda}],
$$
where $\bm{y}(t)$ is \textcolor{black}{known as} the costate variable. 

\begin{lemma}[Pontryagin Maximum Principle \citep{Bittner1963LSP, Sethi2019}]\label{th:pontryagin} If the feasible control ${\bm{u}^*}$ and the state trajectory ${\bm{x}^*}$ is optimal for Problem \eqref{eq:fluid}, there exists ${\bm{y}}(t)$ for any $t \in [0,T]$ that satisfies the following conditions.
\begin{enumerate}
\item[{\bf (a)}]  $H(\bm{x}^*, \bm{u}^*, \bm{y}, t) \leq H(\bm{x}^*, \bm{u}, \bm{y}, t)$ for all $\bm{u}(t)$ satisfying $\bm{u}(t) \geq \bm{0}$, $\bm{D}\bm{u}(t) \leq \bm{e}$. 
\item[{\bf (b)}]  Whenever ${\bm{u}}^*(t)$ is continuous,
    $
    \dot{\bm{y}}(t) =  -\bm{c} + \bm{\pi}(t),
    $
    where $\bm{\pi}(t) \geq \bm{0}, \bm{\pi}(t)^{\top}{\bm{x}^*}(t) = 0$.
\item[{\bf (c)}]  $\bm{y}(T) = \bm{0}$.
\end{enumerate}
\end{lemma}

\begin{proof}
See \citep{Sethi2019}.
\end{proof}

The following two lemmas describe the property of the costate variable in Lemma \ref{th:pontryagin} and the optimal policy, respectively.

\begin{lemma}[\citet{avram1997optimal}]\label{th:index} $\bm{y}(t)$ is piecewise linear and continuous function of $t$. Furthermore, slope changes can only occur at the depletion times $\{T_1, \dots, T_n\}$.
\end{lemma}

\begin{proof}
See \citep{avram1997optimal}.  
\end{proof}

\begin{lemma}[\citet{bauerle2002}]\label{th:scalar}
Consider two initial states $\bm{x}_0$ and $\alpha\bm{x}_0$, where $\alpha$ is some positive scalar. Then, $\bm{u}^{*}_{\bm{x}_0}(0) = \bm{u}^{*}_{\alpha\bm{x}_0}(0)$.
\end{lemma}

\begin{proof}
See \citep{bauerle2002}.
\end{proof}

% \subsection{Robust MFQNETs control}

% In \citep{robustfluid}, a robust version of Problem \eqref{eq:fluid} has been proposed, which immunizes the solution against parameter uncertainties. In this formulation, the arrival rates and the processing rates are  uncertain. They are only assumed to come from known a set, refereed to as the uncertainty set. The resulting solution by solving this robust version gives solutions that are immune against any realization of the uncertain parameters within this uncertainty set. 

\subsection{Robust MFQNETs}
\label{sec:rmfqnets}

\cite{robustfluid} propose a robust formulation for the MFQNET control problems, where the arrival and the service rates are assumed to be uncertain. These parameters belong to a set referred to as the uncertainty set and can take any value within this set (see \cite{Bertsimasrobusttext} for a comprehensive review of robust optimization). 

In their work, they start with a deterministic MFQNET control problem that is slightly different from, but equivalent to Problem \eqref{eq:fluid}. They introduce control variables $\hat{u}_i(t) \triangleq u_i(t)/\tau_i$, $i \in [n]$, which represent the level of effort rather than the fraction of effort for class $i$ fluids. The service time for class $i$ fluids is defined as $\tau_i \triangleq 1/\mu_i$. 

Instead of the binary matrix $\bm{D}$, they define $\tilde{\bm{D}} \in \mathbb{R}^{m \times n}$, where $\tilde{d}_{ji} = {\tau}_i$ if $s(i) = j$ and $\tilde{d}_{ji} = 0$, otherwise. Similarly, instead of using the matrix $\bm{A}$, they use the binary matrix $\hat{\bm{A}} \in \{0,1\}^{n \times n}$, where $\hat{a}_{ii} = 1$ and  $\hat{a}_{ij} = -1$ if class $i$ receives arrivals from class $j, j \neq i$. All its other entries are zero. When there is no uncertainty, this formulation is equivalent to Problem \eqref{eq:fluid}.

In the robust formulation they propose, $\tau_i$ and $\lambda_i$ are now allowed to change over time. Thus, we let $\hat{\tau}_i(t)$ and $\hat{\lambda}_i(t)$ represent the service time and the arrival rate at time $t$, respectively. We also define the $m \times n$ matrix-valued function $\hat{\bm{D}}(\cdot)$, where $\hat{d}_{ji}(t) = \hat{\tau}_i(t)$ if $s(i) = j$ and $\hat{d}_{ji}(t) = 0$, otherwise. For each $t \in [0,T]$, $\hat{\tau}_i(t)$ can take any value in the interval $[\bar{\tau}_i, \bar{\tau}_i + \tilde{\tau}_i]$, where $\bar{\tau}_i$ is the nominal service time and $\tilde{\tau}_i$ is its deviation. We define $z_i(t)$ as the relative deviation from the nominal service time at time $t$, expressed as:
\begin{equation*}
\begin{aligned}
z_i(t) &= 
\begin{cases}
\displaystyle \frac{\hat{\tau_i}(t) - \bar{\tau}_i}{\tilde{\tau}_i} , & \text{if } \tilde{\tau}_i > 0, \\ 
\displaystyle 0, & \text{if } \tilde{\tau}_i = 0. 
\end{cases} 
\end{aligned}
\end{equation*}
The uncertainty set $\mathcal{U}$ is defined as the set of all matrix-valued functions $\hat{\bm{D}}(\cdot)$, where $\hat{\bm{\tau}}(\cdot)$ satisfies the following conditions:
\begin{equation}
\begin{aligned}
    \hat{\tau}_i(t) &= \bar{\tau}_i + z_i(t)\tilde{\tau}_i, && \forall i \in [n], \; t \in [0,T], \\
    \sum_{i:s(i) = j}z_i(t) &\leq \Gamma_j, && \forall j \in [m], \; t \in [0,T], \\
    0 \leq z_i(t) &\leq 1, && \forall i \in [n], \; t \in [0,T].
\end{aligned}
\label{eq:unc_service}
\end{equation}
The parameter $\Gamma_j \in [0, |\{i \in [n] | s(i) = j\}|]$ controls the size of the uncertainty set.

The uncertainties for the arrival rates are defined similarly. At each time $t \in [0,T]$, the time-varying arrival rate $\hat{\lambda}_i(t)$ belongs to the interval $[\bar{\lambda}_i, \bar{\lambda}_i + \tilde{\lambda}_i]$, where $\bar{\lambda}_i$ is the nominal rate and $\tilde{\lambda}_i$ is its deviation. The relative deviation is defined as:
\begin{equation*}
\begin{aligned}
\xi_i(t) &= 
\begin{cases}
\displaystyle \frac{\hat{\lambda}_i(t) - \bar{\lambda}_i}{\tilde{\lambda}_i} , & \text{if } \tilde{\lambda}_i > 0, \\ 
\displaystyle 0, & \text{if } \tilde{\lambda}_i = 0. 
\end{cases} 
\end{aligned}
\end{equation*}
The uncertainty set for the arrival rates is defined as the set of all vector functions $\hat{\lambda}(\cdot)$ that satisfy the following conditions, where $\bm{\Delta}$ is the parameter controlling the size of the uncertainty set:
\begin{equation}
\begin{aligned}
    \hat{\lambda}_i(t) &= \bar{\lambda}_i + \xi_i(t)\tilde{\lambda}_i, && \forall i \in [n], \; t \in [0,T], \\
    \sum_{i:s(i) = j}\xi_i(t) &\leq \Delta_j, && \forall j \in [m], \; t \in [0,T], \\
    0 \leq \xi_i(t) &\leq 1, && \forall i \in [n], \; t \in [0,T].
\end{aligned}
\end{equation}

\cite{robustfluid} show that the robust MFQNET control problem under the described uncertainties can be formulated as the following SCLP problem:
\begin{alignat}{2}
&\underset{\hat{\bm{u}}(\cdot), \bm{x}(\cdot)}\min\quad && \int_{0}^{T} \bm{c}^{\top}\bm{x}(t) \,dt  \label{eq:robfluid}\\
&s.t. && \dot{\bm{x}}(t) = -\hat{\bm{A}}\hat{\bm{u}}(t) + \bar{\bm{\lambda}}, \quad \forall t \in [0,T], \nonumber\\
&\qquad \ && \Gamma_j \beta_j(t) + \sum_{i:s(i) = j}(\bar{\tau}_i\hat{u_i}(t) + \alpha_i(t)) \leq 1, \quad \forall j \in [m], t \in [0,T],  \nonumber \\
&\qquad \ && \alpha_i(t) + \beta_j(t) - \hat{u_i}(t)\tilde{\tau}_i \geq \bm{0}, \quad \forall j,i \; \text{with} \; s(i)=j, t \in [0,T], \nonumber \\
&\qquad \ && \hat{\bm{u}}(t), \bm{x}(t), \bm{\alpha}(t), \bm{\beta}(t) \geq \bm{0}, \quad \forall t \in [0,T], \nonumber \\
&\qquad \ && \bm{x}(0) = \bm{x}_0. \nonumber
\end{alignat}
Note that Problem \eqref{eq:robfluid} does not depend on the uncertainties in the arrival rates. Therefore, we assume the arrival rates are deterministic without loss of generality. For the remainder of the paper, we refer to MFQNETs with uncertain service times (or equivalently, uncertain service rates) as robust MFQNETs. MFQNETs without uncertainty (as described in Section \ref{sec:fluid}) are referred to as deterministic MFQNETs, or simply MFQNETs. \cite{robustfluid} demonstrate that the policy derived from robust MFQNETs achieves strong empirical performance in controlling the underlying stochastic networks.

\subsection{Optimal Classification Trees with Hyperplane Splits}
\label{sec:oct}

Optimal Classification Trees (OCT) is an algorithm to learn near-optimal decision trees for classification tasks. Classification and Regression Trees (CART) \citep{BreiFrieStonOlsh84}, an earlier algorithm to learn decision trees for prediction tasks, learns a decision tree in a greedy manner using recursive partitioning of the feature space at each child node. However, OCT aims to learn a globally optimal decision tree using mixed\textcolor{black}{-}integer optimization and local heuristics. 

Similar to CART and other classification algorithms, OCT takes $N$ data inputs $\{(\bm{\theta}_i, z_i)\}_{i=1}^{N}$, where $\bm{\theta}_i$ is the feature vector and $z_i$ is the label for the $i_{th}$ data point. Given this data set, OCT learns a decision tree that uses a single feature for the split at each node and assigns a label to each node of the tree. Given a new data point $\bm{\theta}_0$, it traverses the decision tree until it reaches a leaf node. The prediction of the tree for $\bm{\theta}_0$ is the label assigned to the leaf node. 

OCT-H, a generalization of OCT, can use an arbitrary linear combination of the features for splits at the nodes. This means that OCT-H can use general hyperplanes for splits, whereas OCT is confined to use hyperplanes that are perpendicular to the axes in the feature space. Essentially, OCT-H partitions the feature space with hyperplanes, and assigns a prediction to each region. This observation is the key to our work to solve Problem \eqref{eq:fluid} using OCT-H. Compared to OCT, OCT-H generally shows higher prediction accuracy and learns shallower trees. 

In OCT-H, it is possible to limit the number of features that can be used for splits, which can result in a more interpretable tree. This version of OCT-H is denoted as OCT-H with sparsity throughout the remainder of the paper. We simply use OCT-H to denote regular OCT-H, where the entire features can be used. For a more detailed explanation on OCT and OCT-H, we refer readers to \citet{OCT, MLOPT}.

\section{OCT-H for the Optimal Control of MFQNETs}
\label{sec:main}

In this section, we first prove that OCT-H can learn an optimal policy of Problem \eqref{eq:fluid}. Based on this result, we then proceed to develop an efficient algorithm to learn an optimal policy of Problem \eqref{eq:fluid} using OCT-H. Finally, we demonstrate that the theoretical results and the proposed algorithm also extend to robust MFQNETs.

\subsection{Theoretical Results}
\label{sec:theory}

The following Lemma \ref{th:costate} will be used to prove our main theorem.

\begin{lemma}\label{th:costate} The value of $\bm{y}(0)$ can be expressed as a linear function of $(T_1, \dots, T_n)$.
\end{lemma}

\begin{proof}

\noindent
We fix an index $i \in [n]$ and prove the stated property for $y_i(0)$. By Lemma \ref{th:index}, $y_i(t)$ is piecewise linear and continuous, with potential breakpoints at the depletion times $\{T_1, \dots, T_n\}$. Although the exact ordering of the depletion times is unknown, the following argument applies to any ordering. Hence, assume a fixed ordering. By working backwards in time from $t = T$ (where $y_i(T) = 0$ by Lemma \ref{th:pontryagin}), we can determine the value of $y(t)$ at the breakpoints $\{T_1, \dots, T_n\}$, starting from the breakpoint that is closest to $T$. At any breakpoint, the value of $y_i(t)$ can be expressed as a linear combination of $T_1, \dots, T_n$ that are greater than or equal to $t$. Once the value of $y_i(t)$ at a breakpoint is determined, the value of $y_i(t)$ at the next earlier breakpoint can be determined, as we know that $y_i(t)$ is piecewise linear and continuous.  We recursively follow this procedure until $t = 0$, where the value of $y_i(0)$ can be expressed as a linear combination of $(T_1, \dots, T_n)$. 
\end{proof}

The following theorem is our main theoretical result that generalizes the results by \citet{avrambertsimasricard} to general MFQNETs.

\begin{theorem}\label{th:main}
For Problem \eqref{eq:fluid}, there exists a piecewise constant optimal policy, with segments separated by hyperplanes passing through the origin.
\end{theorem}

\begin{proof}

Our proof is based upon the algorithm by \citet{avram1997optimal} to solve Problem \eqref{eq:fluid} with any given initial state. We prove that the solution this algorithm finds follows a piecewise constant optimal policy, with segments separated by hyperplanes passing through the origin.

 By condition (a) of Lemma \ref{th:pontryagin}, the optimal control at each time $t$ is decided by the priority index $r_{i}(t)$ defined for each fluid class $i \in [n]$, where $r_{i}(t) = [\bm{y}(t)^{\top}\bm{A}]_{i}$. Lemma \ref{th:index} indicates that $r_{i}(t)$ is a continuous, piecewise linear function and its slope can only change at $\{T_1, \dots, T_n\}$.  At each server, the optimal policy is to put maximum effort to the fluid class with the smallest priority index, and put zero effort to the rest, without violating the non-negativity constraint on the state variables. When all the fluid classes at the server have positive priority indices, the optimal policy is to idle. This case can be captured by considering idling as a fluid class with the constant priority index 0. Hence, Lemma \ref{th:pontryagin} implies that as long as the rank of the priority indices does not change, the optimal control is a constant vector. 

The condition under which a server transfers effort from one class to another is defined by the equalities of the form $r_{j}(t) = r_{k}(t)$, given that class $j$ fluids and class $k$ fluids are processed by the same server. If this equality holds, then it is indifferent whether the server prioritizes class \textcolor{black}{$j$} or $k$. If this equality becomes inequality, then it would be beneficial to prioritize one class over the other. This description indicates that the switching between fluid classes are defined by the equalities between the priority indices. 

We derive the condition that the priority is switched from class $j$ fluids to class $k$ fluids, starting from an initial state $\bm{x}_0$. Depending on the parameters $\bm{c}$ and $\bm{A}$, certain switches might not be always possible. Furthermore, the order of the depletion times $\{T_1, \dots, T_n\}$ and the future switches associated with the trajectory should be adequately decided as well (Specific examples of how $\bm{A}$, $\bm{c}$ and the order of $\{T_1, \dots, T_n\}$ can make a switch possible or not are given in \citep{avrambertsimasricard, avram1997optimal}). We assume that $\bm{A}$, $\bm{c}$, the order of $\{T_1, \dots, T_n\}$ and the switches associated with the trajectory are appropriately fixed. The specific order of $\{T_1, \dots, T_n\}$ leads to a collection of equalities between the indices $r_i(t)$
during the entire trajectory, and completely determines the optimal solution of the problem \citep{avrambertsimasricard, avram1997optimal}. 

Without loss of generality, we assume that the switch from class $j$ to $k$ happens at $t = 0$. The switching condition that we would like to derive is then $r_j(0) = r_k(0)$, where $r_j(0)$ and $r_k(0)$ 
 are both linear functions of $(T_1, \dots, T_n)$ due to Lemma \ref{th:costate}. We now prove that  $(T_1, \dots, T_n)$ is a linear function of $\bm{x}_0$, which verifies that $r_j(0) = r_k(0)$ represents a hyperplane passing through the origin in the state space.

 By definition, $T_i$ can be computed from the equation of the form $\int_{0}^{b_1} [\bm{A}\bm{u}(t) + \bm{\lambda}]_i  \,dt + \dots + \int_{b_{q}}^{T_i} [\bm{A}\bm{u}(t) + \bm{\lambda}]_i \,dt = -x_i$, where \textcolor{black}{$b_l, l \in [q] $}, represents a breakpoint in $[\bm{A}\bm{u}(t) + \bm{\lambda}]_i $ . The control $[\bm{A}\bm{u}(t) + \bm{\lambda}]_i$ is constant between the breakpoints, and the breakpoints $b_i$ are always the intersections between two indices. Any time of intersection between two indices can be expressed as a linear function of the vector $(T_1, \dots, T_n)$ due to Lemma \ref{th:costate}. Hence, the above equation leads to an equality between $x_i$ and a linear function of $(T_1, \dots, T_n)$. Likewise, the collection of equalities that we have all \textcolor{black}{lead} to equalities between $\bm{x}_0$ and linear functions of $(T_1, \dots, T_n)$. Rearranging these equalities leads to the expression of $(T_1, \dots, T_n)$ as a linear function of $\bm{x}_0$. 
\end{proof}

The proof above demonstrates that the state space is separated by hyperplanes, according to the relative ordering of the priority indices of each fluid class. Within each region, a constant control vector is optimal. These hyperplanes are referred to as switching curves \citep{Meyn2007, Sethi2019}.

The following Corollary \ref{th:oct} is the building block to develop a learning algorithm in Section \ref{sec:algorithm}.
\begin{corollary}
\label{th:oct}
OCT-H can learn an optimal policy of Problem \eqref{eq:fluid}.
\end{corollary}

\begin{proof}

By Theorem \ref{th:main}, there exist switching curves that are hyperplanes passing through the origin. As described in Section \ref{sec:oct}, OCT-H learns a decision tree that partitions the feature space with hyperplanes, and assigns a label to each region. Hence, it can naturally learn the switching curves and the optimal control at each region partitioned by the switching curves. 

Another condition to consider is whether $x_i(t) = 0$ for some class $i \in [n]$. If $x_i = 0$, then server splitting might occur to satisfy the non-negativity constraint on the state vector. We first note that the condition $x_i = 0$ is also a hyperplane in the state space passing through the origin. 

In general, decision trees are confined to use inequalities for node splits. In our context, however, equality conditions such as $x_i(t) = 0$ can be learned as the condition $x_i \leq 0$. Since the state vector is always non-negative, these two conditions are equivalent for Problem \eqref{eq:fluid}.
Hence, OCT-H can learn the optimal policy of Problem \eqref{eq:fluid} both in the interior and the boundary of the state space. 
\end{proof}

The following \textcolor{black}{Proposition \ref{th:scalar2}}, along with Lemma \ref{th:scalar} will be used in Section \ref{sec:algorithm} to develop a more efficient learning algorithm.

\color{black}

\begin{proposition}\label{th:scalar2}
Consider a pair of optimal control and the associated state trajectory $\{(\bm{u}_{\bm{x}_0}^*(t), \bm{x}^*_{\bm{x}_0}(t)): t \in [0,T]\}$ and a positive scalar $\alpha$. Then, $\{(\bm{u}_{\bm{x}_0}^*(\frac{t}{\alpha}), \alpha \bm{x}^*_{\bm{x}_0}(\frac{t}{\alpha})): t \in [0,\alpha T]\}$ is optimal for Problem \eqref{eq:fluid} with the initial state  $\alpha \bm{x}_{0}$.
\end{proposition}

\begin{proof}
The proof of this theorem follows  from the proof of Theorem 3 in \citep{bauerle2002}. By Theorem 3 in  \citep{bauerle2002}, we know $V(\alpha \bm{x}_0) = \alpha^2 V(\bm{x}_0)$ and also that $\{(\bm{u}_{\bm{x}_0}^*(\frac{t}{\alpha}), \alpha \bm{x}^*_{\bm{x}_0}(\frac{t}{\alpha})): t \in [0,\alpha T]\}$ is feasible. The objective cost associated with the pair $\{(\bm{u}_{\bm{x}_0}^*(\frac{t}{\alpha}), \alpha \bm{x}^*_{\bm{x}_0}(\frac{t}{\alpha})): t \in [0,\alpha T]\}$ is 
$$
\alpha \int_{0}^{\alpha T} \bm{c}^{\top}\bm{x}^*_{\bm{x}_0}(\frac{t}{\alpha}) \,dt  
= \alpha^2 \int_{0}^{T} \bm{c}^{\top}\bm{x}^*_{\bm{x}_0}(t) \,dt
= \alpha^2 V(\bm{x}_0).
$$
As this solution achieves the optimal objective cost and is also feasible, it is optimal. 
\end{proof}

\color{black}

\subsection{Algorithm}
\label{sec:algorithm}

 \textcolor{black}{We present an algorithm that utilizes OCT-H to learn an optimal policy for Problem \eqref{eq:fluid}. To ensure a more comprehensive and efficient learning process, we discuss several key considerations that have been taken into account while developing the algorithm.}  

Given Problem \eqref{eq:fluid} with the initial state $\bm{x}_0$, we can solve it to optimality using the algorithm proposed by \citet{evgenyweiss2021}. Once we solve it, we \textcolor{black}{obtain} the optimal control $\bm{u}_{\bm{x}_0}^*(t)$ and the optimal state trajectory $\bm{x}_{\bm{x}_0}^{*}(t)$ for the entire time interval $t \in [0,T]$. We choose $N \in \mathbb{N}$ elements $t_1, \dots, t_N$ from the interval $[0,T]$, and extract the corresponding state values $\bm{x}^*(t_1), \dots, \bm{x}^*(t_N)$ and the control values $\bm{u}^*(t_1), \dots, \bm{u}^*(t_N)$. The training data that we obtain from this procedure is $\{(\bm{x}_{\bm{x}_0}^*(t_i), \bm{u}_{\bm{x}_0}^*(t_i))\}_{i=1}^{N}$.  As in the usual supervised learning literature, $\bm{x}_{\bm{x}_0}^*(t_i)$ is the feature vector and $\bm{u}_{\bm{x}_0}^*(t_i)$ is the target for the $i_{th}$ data point. To ensure a comprehensive coverage of the state space, we generate multiple initial states and solve the associated Problem \eqref{eq:fluid} for each initial state. 

{\color{black}For scenarios in which one seeks to train a policy that covers the entire state space $\mathbb{R}^n_{\ge 0}$, we propose a systematic approach for generating initial states.} Assuming that there are $n$ fluid classes, there are $\binom{n}{1} + \dots + \binom{n}{n} = 2^{n} - 1$ possible cases of which classes among $n$ are non-empty (excluding the trivial case that the entire system is empty). We let $\mathcal{S} = \{s_1, s_2, \dots, s_{2^n - 1} \}$ be the set of such cases, where each element $s_i, i \in [2^n-1], $ represents a set of non-empty classes. For example, if $n=2$, then $\mathcal{S} = \Big\{\{1\},\{2\}, \{1,2\} \Big\}$. For each $s \in \mathcal{S}$, we generate values for the non-zero entries of the initial state specified in $s$ and fix the remaining entries to zero. This systematic approach ensures that the training data covers the state space seamlessly, including both the interior and the boundary regions. 

Another consideration is that as $n$ increases, the number of hyperplanes required to describe the optimal policy can get prohibitively large. To address this issue, we propose training multiple decision trees, if necessary. Each data point in the training set corresponds to an element of $\mathcal{S}$, depending on which entries of the state vector are non-zero. Hence, once we define a partition of $\mathcal{S}$, this partition can also be used to partition the training set. Then, we train a decision tree for each partition of the training set. For example, if we define a partition of the set $\mathcal{S} = \Big\{\{1\},\{2\}, \{1,2\} \Big\}$ to be  $\mathcal{P} = \Bigg\{ \Big\{ \{1\},\{2\}    \Big\} ,\Big\{\{1,2\} \Big\}    \Bigg\}$, we train two decision trees. The first decision tree is trained using the state vectors where either the first or the second entry is zero. The second decision tree is trained using the state vectors that are strictly positive. Essentially, we are dividing the state space into multiple regions and learning the optimal policy for each region. {\color{black} At test time, given a new state vector, we first determine which partition it belongs to and then apply the corresponding policy.} 

{\color{black} In our setting, combining these locally trained decision-tree policies can yield a globally optimal policy defined over the entire state space. By Theorem \ref{th:main}, this way of partitioning ensures that the switching curve within each region is completely independent of the switching curves in other regions. As a result, each local policy can be learned perfectly from data restricted to its own region, without requiring samples from the rest of the state space.} This approach allows us to distribute the learning process across multiple decision trees, reducing the computational burden and enabling efficient training even when dealing with a large number of fluid classes.

Finally, we use Lemma \ref{th:scalar} and Proposition \ref{th:scalar2} for a more efficient data generation. According to Proposition \ref{th:scalar2}, solving Problem \eqref{eq:fluid} with the initial state $\bm{x}_0$ and solving it again with $\alpha \bm{x}_0$ would be redundant. Hence, instead of generating initial states arbitrarily, we can sample them only from the unit sphere in the non-negative orthant. Furthermore, Lemma \ref{th:scalar} suggests that we can augment the training data $\{(\bm{x}_{\bm{x}_0}^*(t_i), \bm{u}_{\bm{x}_0}^*(t_i))\}_{i=1}^{N}$ by including additional data points $\{(\alpha\bm{x}_{\bm{x}_0}^*(t_i), \bm{u}_{\bm{x}_0}^*(t_i))\}_{i=1}^{N}$, possibly multiple times with varying $\alpha$ values.

Algorithm \ref{alg:main} outlines the entire procedure more rigorously. We let $\mathcal{A}$ denote the set of $\alpha$ that we use to augment data. We let $\mathcal{P}$ denote the partition of  $\mathcal{S}$. We use $M$ to denote the number of initial states we sample for each element in $\mathcal{S}$. We use $\bm{x}_{[s]}$ to denote the entries of $\bm{x}$ in $s$. For a set $K = \{(\bm{x}_{\bm{x}_0}^*(t_i), \bm{u}_{\bm{x}_0}^*(t_i))\}_{i=1}^{N}$, we use $\alpha K$ to denote $\{(\alpha\bm{x}_{\bm{x}_0}^*(t_i), \bm{u}_{\bm{x}_0}^*(t_i))\}_{i=1}^{N}$. Without loss of generality, we assume that the order of the cells in the partition $\mathcal{P}$ is fixed and $\mathcal{P}_{[j]}$ is the $j_{th}$ cell of $\mathcal{P}$.

{\color{black}
\begin{remark} \label{remark:dist}  In practice, learning a policy over the entire state space $\mathbb{R}^n_{\ge 0}$ may be an overly ambitious goal, depending on the available computational resources. A more realistic scenario could be that one is interested in a specific subset of the state space or assumes a particular distribution over states, assumptions that underlie virtually all statistical learning frameworks. In such settings, it is unnecessary to generate training data as exhaustively as described above. Instead, one can generate initial states within the region of interest or sample them from the distribution of interest and construct training data accordingly. The remainder of the procedure can then be followed exactly as described, including partitioning the training data and training separate policies for each partition. 
\end{remark}}

\begin{algorithm}
 \KwInput{$\bm{c}, \bm{A}, \bm{\lambda},\bm{D}, \{t_1, \dots, t_N\}, \mathcal{A}, \mathcal{S}, \mathcal{P}, M$}
\KwOutput{$|\mathcal{P}|$ classification trees with hyperplane splits.} 
\textbf{Initialization: $K_{\mathcal{S}}, K_1, \dots, K_{|\mathcal{P}|} \leftarrow \emptyset$} 

\vspace{3mm}

\textbf{1. Data Generation} \\
%  \For{$j \in [|\mathcal{P}|]$}{
%  $p \leftarrow \mathcal{P}_{[j]}$ \\
%   \For{$s \in p$}{
%   \For{$i \in [M]$}{
%     $\bm{x}_0 \leftarrow \bm{0} \in \mathbb{R}^n$\\
%     Sample a positive vector $\hat{\bm{x}}$ from the $|s|$ dimensional unit sphere. \\
%     $\bm{x}_{0[s]} \leftarrow \hat{\bm{x}}$ \\
%     Solve problem \eqref{eq:fluid} with the initial state $\bm{x}_0$, denote the optimal control and state trajectory as $\bm{u}^*(\cdot), \bm{x}^*(\cdot)$. \\
%     $K_j \leftarrow K_j \cup \{(\bm{x}_{\bm{x}_0}^*(t_i), \bm{u}_{\bm{x}_0}^*(t_i))\}_{i=1}^{N}$
% }
%   }

%   }
\For{$s \in \mathcal{S}$}{
  $j \leftarrow 1$ \\
  \While{$j \leq M$}{
    $\bm{x}_0 \leftarrow \bm{0} \in \mathbb{R}^n$\\
    Sample a positive vector $\hat{\bm{x}}$ from the $|s|$ dimensional unit sphere. \\
    $\bm{x}_{0[s]} \leftarrow \hat{\bm{x}}$ \\
    Solve Problem \eqref{eq:fluid} with the initial state $\bm{x}_0$. \\
    $K_{\mathcal{S}} \leftarrow K_{\mathcal{S}} \cup \{(\bm{x}_{\bm{x}_0}^*(t_i), \bm{u}_{\bm{x}_0}^*(t_i))\}_{i=1}^{N}$ \\
    $j \leftarrow j+1$
}
  }

\For{$(\bm{x}, \bm{u}) \in K_{\mathcal{S}}$}{
 $\hat{s} \leftarrow \{i \in [n]: x_i > 0\}$ \\

 \For{$j \in [|\mathcal{P}|]$}{
\For{$s \in \mathcal{P}_{[j]}$}{
\If{$s = \hat{s}$}{
${K}_j \leftarrow {K}_j \cup (\bm{x}, \bm{u})$
}

}
 }

}

  \vspace{3mm}

\textbf{2. Data Augmentation} \\
\For{$K \in \{K_1, \dots, K_{|\mathcal{P}|}\}$}{
  \For{$\alpha \in \mathcal{A}$}{
    $K \leftarrow K \cup \alpha K$
  }
  }

  \vspace{3mm}

\textbf{3. Training} \\
\For{$K \in \{K_1, \dots, K_{|\mathcal{P}|}\}$}{
Use OCT-H to train a classification tree on $K$.
}

  \caption{OCT-H for MFQNET control.}
 \label{alg:main}
\end{algorithm}

\subsection{Example}
\label{sec:example}

We provide two small examples to illustrate Algorithm \ref{alg:main}. For both examples, the closed-form expressions of the optimal policies are already known. We compare the policy learned by Algorithm \ref{alg:main} with the closed-form optimal policy to demonstrate that it can learn near-optimal policies. The first example is to demonstrate that Algorithm \ref{alg:main} can learn the optimal switching curve in the interior of the state space, and the second example is to demonstrate that it can learn the optimal server splitting policy when some fluid classes are empty. In Section \ref{sec:exp_rob}, we apply these policies to multiple test instances and demonstrate that the resulting objective costs are near-optimal.

\begin{figure}
    \centering
    \includegraphics[scale = 0.35]{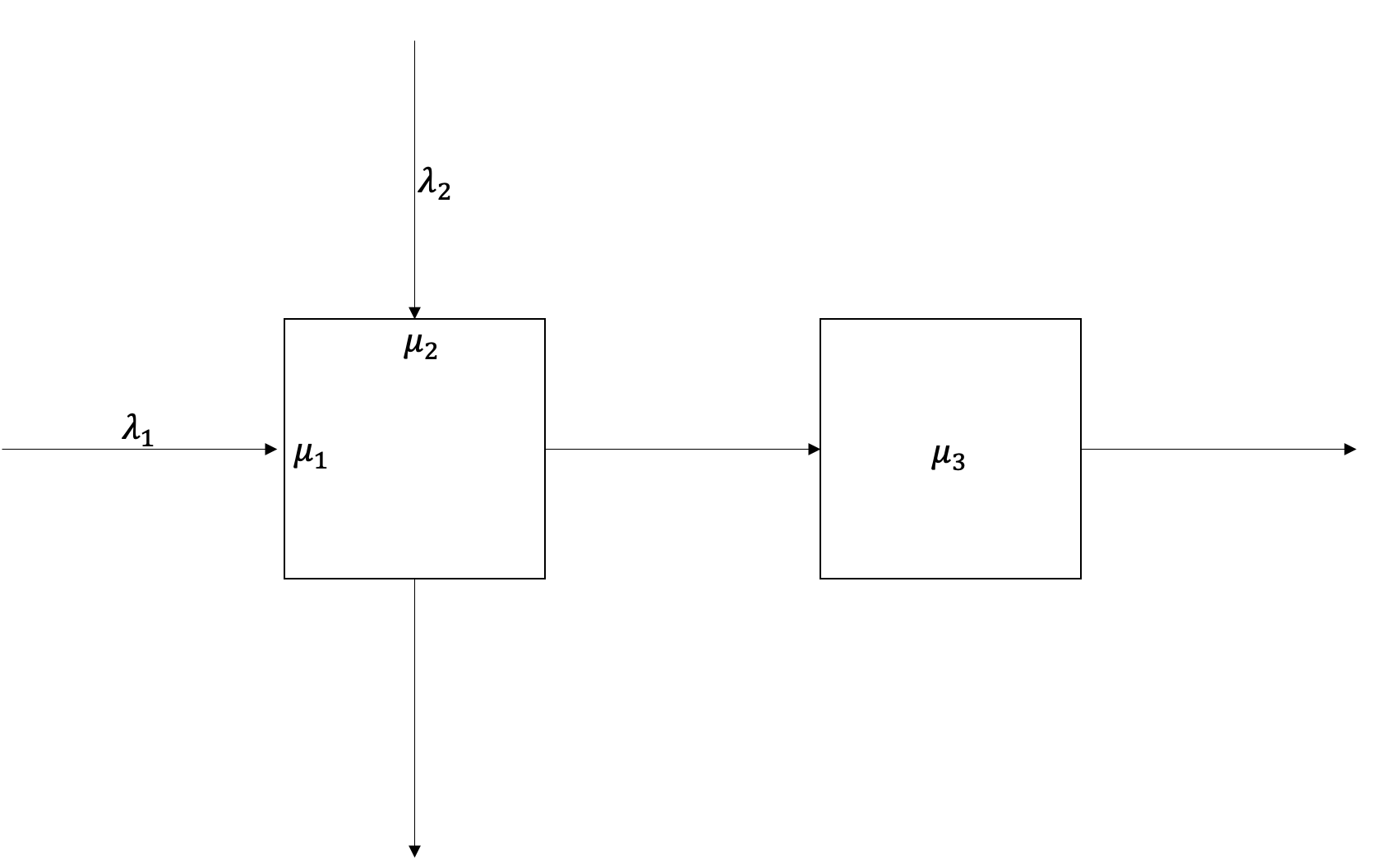}
    \caption{Criss-cross network.}
    \label{fig:crisscross}
\end{figure}

The first example is the criss-cross network considered by \citet{crisscross}. The criss-cross network is composed of three classes and two servers. Server 1 processes Class 1 and 2 fluids, and Server 2 processes Class 3 fluids. Class 1 and 2 fluids take external arrivals. After Class 1 fluids are processed at Server 1, they become Class 3 fluids and move to Server 2. After Class 2 and 3 fluids are processed, they leave the system. Its graphical representation is given in Figure \ref{fig:crisscross}. It is clear that $u_3^*(t) = 1$ as long as Server 2 is not empty. The problem is to choose which class to process at Server 1. For this example, we only demonstrate the case in which none of the classes are empty. In other words, we learn the optimal policy for the case $s = \{1,2,3\}$. We let $\bm{c} = \bm{e}$, $\lambda_1 = \lambda_2 = 0.5$, $\mu_1 = 1.5$, $\mu_2 = 1$ and $\mu_3 = 2$. Under this set of parameters, the switching curve and the corresponding optimal policy of this network derived by \citet{avrambertsimasricard} are given in Table \ref{table:crisscross}. The closed-form expression of the switching curve is $x_1(t) = 6x_3(t)$. Other parameters we used for Algorithm \ref{alg:main} are $N = 1, t_1 = 0, \mathcal{A} = \{0.5, 1.5\}, \mathcal{P} = \bigg\{\big\{\{s_1,s_2,s_3\}\big\}, \dots \bigg\}$ and $ M = 1000$.

% \begin{table*}\centering
% \ra{1.3}

% \begin{tabular}{lll}
% \toprule
%   Case & Conditions  & $\bm{u}(t)$    \\ 
% \midrule
% Case 1 & $c_1\mu_1 \leq c_3\mu_3$ & (0,1,1)\\ 
% \hline
% \multirow{2}{*}{Case 2} & $c_1\mu_1 \geq c_3\mu_3$& \multirow{2}{*}{(1,1,0)}\\ 
% &$c_1\mu_1 - c_3\mu_3 \geq  c_2\mu_1$& \\
% \hline
% \multirow{3}{*}{Case 3} & $c_1\mu_1 \geq c_3\mu_3$ & \multirow{3}{*}{(0,1,1)}\\
%  &  $c_1\mu_1 - c_3\mu_3 \leq  c_2\mu_1$ & \\
%   &   $\mu_1 \geq \mu_2$ & \\

% \hline 
% \multirow{4}{*}{Case 4} & $c_1\mu_1 \geq c_3\mu_3$ & \multirow{4}{*}{(1,1,0)}\\\
%  & $c_1\mu_1 - c_3\mu_3 \leq  c_2\mu_1$ & \\
%   &$\mu_1 \leq \mu_2$& \\
%    &$\frac{x_1(t)}{x_2(t)} \geq \frac{c_2\mu_1}{c_1\mu_1 - c_3\mu_3} \times \frac{\mu_1 - \lambda_1}{\mu_2 - \mu_1}$& \\

% \hline
% \multirow{4}{*}{Case 5} & $c_1\mu_1 \geq c_3\mu_3$ & \multirow{4}{*}{(0,1,1)}\\ 
%  & $c_1\mu_1 - c_3\mu_3 \leq  c_2\mu_1$ & \\
%   &$\mu_1 \leq \mu_2$& \\
%    &$\frac{x_1(t)}{x_2(t)} \leq \frac{c_2\mu_1}{c_1\mu_1 - c_3\mu_3} \times \frac{\mu_1 - \lambda_1}{\mu_2 - \mu_1}$& \\

% \bottomrule
% \end{tabular}
% \caption{Optimal policy for the criss-cross network.}
% \label{table:crisscross}
% \end{table*}

\begin{table*}\centering

\begin{tabular}{cc}
\toprule
Conditions  & $\bm{u}^*(t)$    \\ 
\midrule

$\frac{x_1(t)}{x_3(t)} \geq \frac{c_2\mu_1}{c_1\mu_1 - c_3\mu_3} \times \frac{\mu_1 - \lambda_1}{\mu_2 - \mu_1}$& $(1,0,1)$\\

\hline

$\frac{x_1(t)}{x_3(t)} \leq \frac{c_2\mu_1}{c_1\mu_1 - c_3\mu_3} \times \frac{\mu_1 - \lambda_1}{\mu_2 - \mu_1}$& $(0,1,1)$\\

\bottomrule
\end{tabular}
\caption{Optimal policy for the criss-cross network when $\bm{x}(t) > \bm{0}$.}
\label{table:crisscross}
\end{table*}

\begin{figure}
    \centering
    \includegraphics[scale = 0.7]{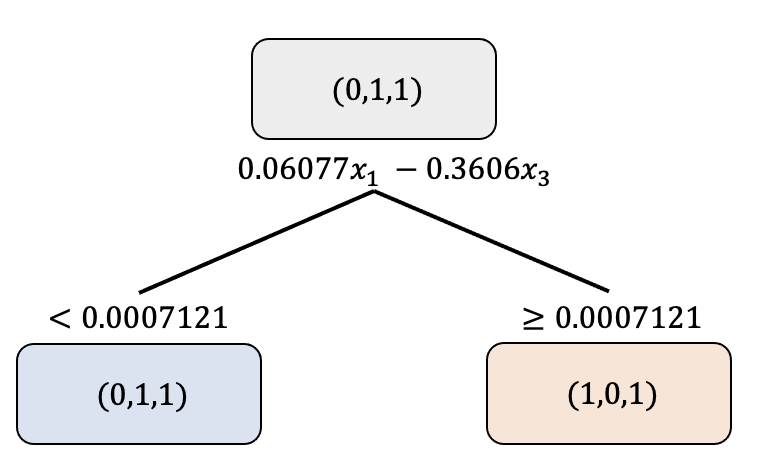}
    \caption{The decision tree learned by OCT-H for the criss-cross network.}
    \label{fig:crisscross_tree}
\end{figure}

Figure \ref{fig:crisscross_tree} displays the decision tree learned by OCT-H, where each node contains the prediction made on that node. The decision tree that OCT-H learned predicts $\bm{u}^*(t) = (0,1,1)$ if $x_1(t)  \leq 5.93x_3(t) + 0.01$ and predicts $\bm{u}^*(t) = (1,0,1)$ if $x_1(t)  \geq 5.93x_3(t) + 0.01$. This closely resembles the optimal policy, exhibiting only minor numerical differences. 

\begin{figure}
    \centering
    \includegraphics[scale = 0.35]{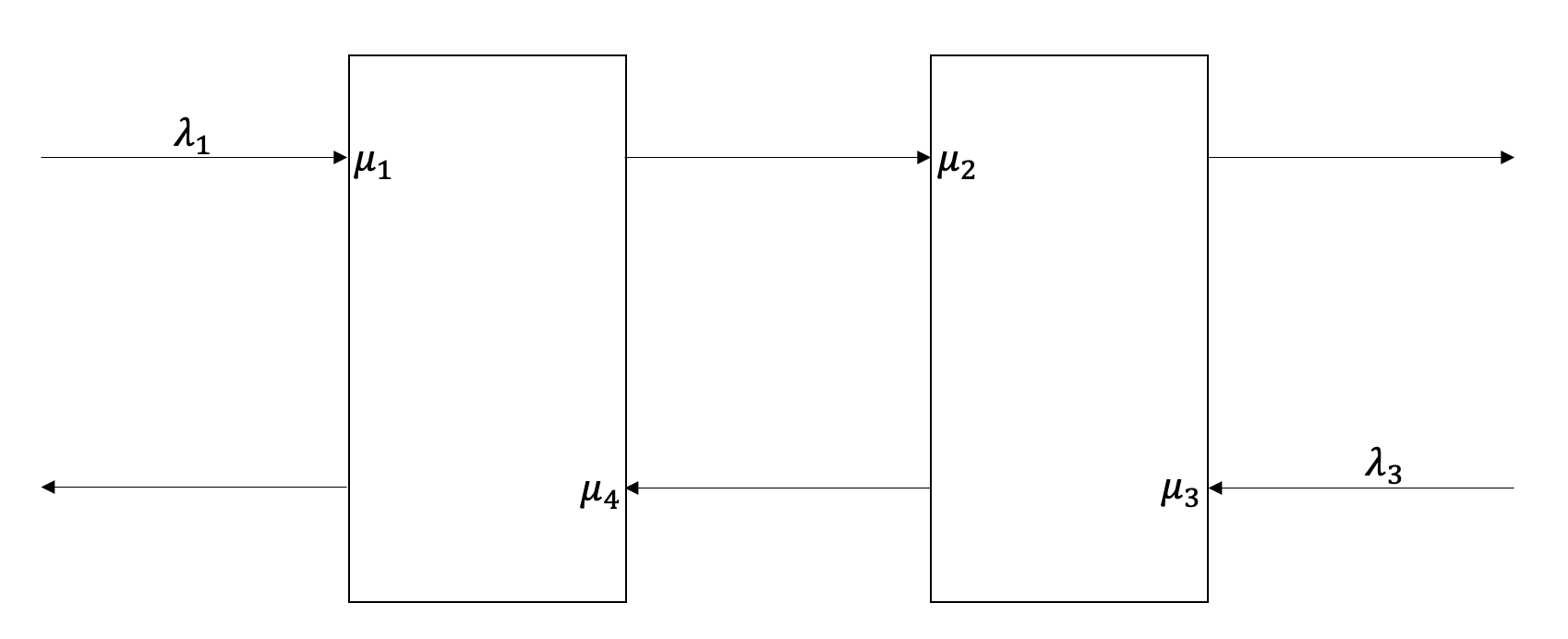}
    \caption{Rybko-Stolyar network.}
    \label{fig:rybko}
\end{figure}

The second example is the Rybko-Stolyar network studied in \citet{RybkoStolyar92}. This network is composed of four classes and two servers. Server 1 processes Class 1 and 4, and Server 2 processes Class 2 and 3 fluids. Class 1 and 3 fluids take external arrivals. After Class 1 fluids are processed, they become Class 2 fluid and move to Server 2. After Class 3 fluids are processed, they become Class 4 fluids and move to Server 1. After Class 2 and 4 fluids are processed, they exit the system. Its graphical representation is given in Figure \ref{fig:rybko}. We let $\bm{c} = \bm{e}$, $\lambda_1 = \lambda_3 = 1$, $\mu_1 = \mu_3 = 6$ and $\mu_2 = \mu_4 = 1.5$. Under this set of parameters, the optimal policy is to prioritize Class 2 and 4 fluids unless either one of them is empty. If any one of them is empty, server splitting occurs. For this example, we train a single decision tree to learn the optimal policy that covers the entire state space. The parameters we used for Algorithm \ref{alg:main} are $N = 1, t_1 = 0, \mathcal{A} = \{0.5, 1.5\}, \mathcal{P} = \{\mathcal{S} \}$ and $ M = 1000$. 

\begin{figure}
\centering
    \includegraphics[scale = 0.5]{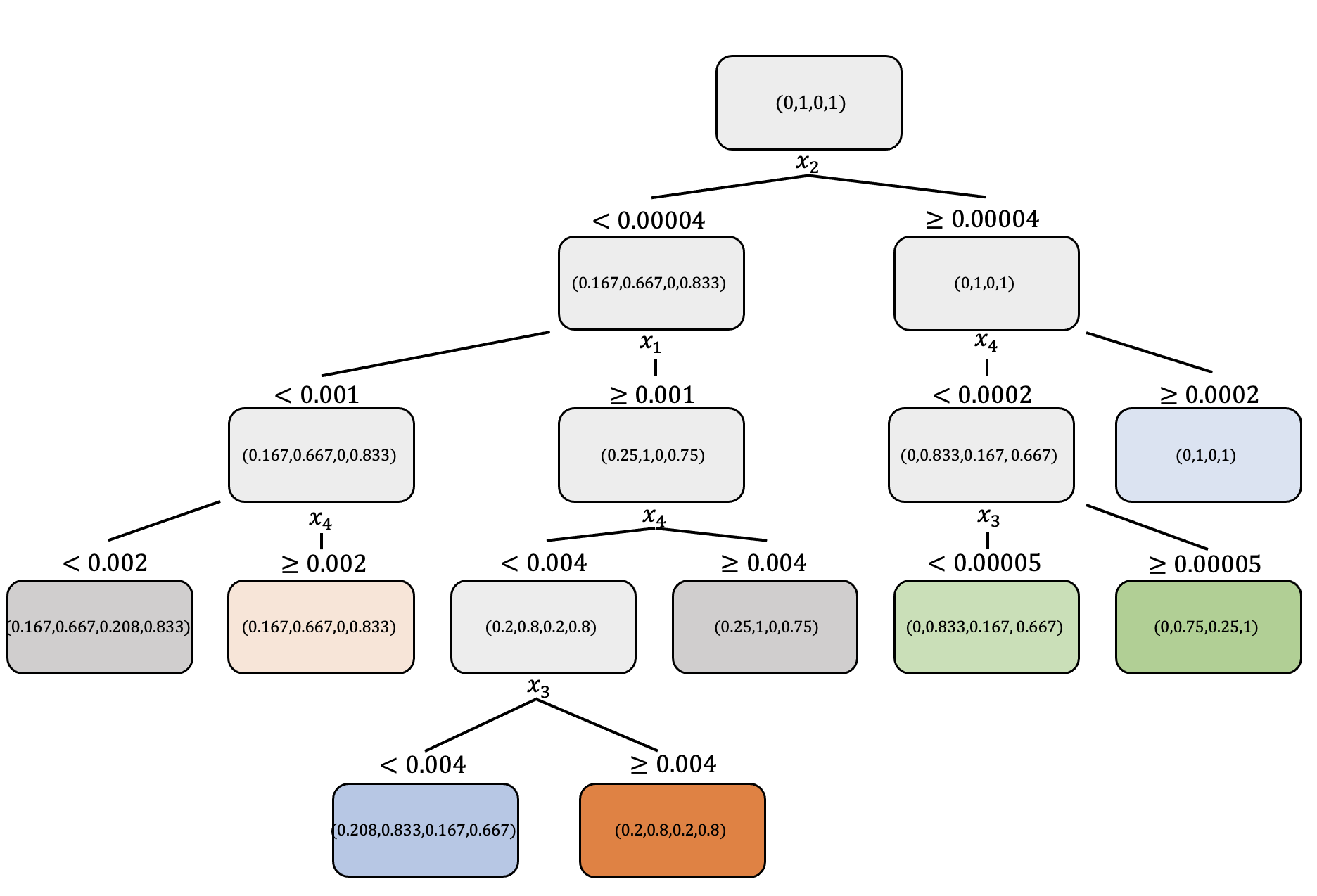}
    \caption{The decision tree learned by OCT-H for the Rybko-Stolyar network.}
    \label{fig:rs_tree}
\end{figure}

Figure \ref{fig:rs_tree} displays the decision tree learned by OCT-H. Since decision trees are confined to use inequalities for node splits, we can observe that the condition $x_i = 0$ for some $i \in [4]$ is learned as $x_i \leq \epsilon$ for a number $\epsilon$ with small absolute value. As the state vectors are always non-negative, these two conditions are effectively equivalent. Additionally, when both Class 2 and 4 are non-empty, the decision tree prioritizes them. When either one of them is empty, server splitting occurs. This policy aligns with the description provided earlier. 

% \begin{figure}
%     \centering
%     \includegraphics[scale = 0.15, angle = 270]{Fluid/image/RS.png}
%     \caption{The decision tree learned by OCT-H for the Rybko-Stolyar.}
%     \label{fig:rs_tree}
% \end{figure}

\subsection{OCT-H for Robust MFQNETs}

In practice, problem parameters may change frequently. In such cases, frequent retraining of policies may become necessary because a learned policy might lose its optimality with respect to the new problem, potentially leading to suboptimal performance. Furthermore, arrival and service rates are often subject to uncertainty, making it difficult to specify these parameter values precisely. To address these challenges, we propose using the optimal policy for robust MFQNETs. Since robust MFQNETs are designed to handle parameter uncertainties, their optimal policies are expected to be more resilient to changes or misspecification of parameters. 

We now show that the key results derived in this section apply directly to robust MFQNETs.
\begin{corollary}
\label{th:rob_oct}
Assuming that the fluids in Problem \eqref{eq:robfluid} can be drained in finite time, OCT-H can learn its optimal policy.
\end{corollary}

\begin{proof}
The structures of Problems \eqref{eq:fluid} and \eqref{eq:robfluid} are identical. The only difference lies in the set of feasible controls, where for Problem \eqref{eq:robfluid}, there are additional dummy control variables, $\bm{\alpha}$ and $\bm{\beta}$, along with extra constraints on these variables. However, this feasible control set remains a polyhedron. Therefore, the proof for Theorem \ref{th:main} applies directly to Problem \eqref{eq:robfluid}.  
\end{proof}

The stability assumption in Corollary \ref{th:rob_oct} can be verified by computing the worst-case vector load over the uncertainty set. Specifically, we check if the following condition holds:
\[
\max_{\hat{\bm{D}}(\cdot) \in \mathcal{U},\, t \in [0,T],\, j \in [m]} \left( -\left[ \hat{\bm{D}}(t)\hat{\bm{A}}^{-1} \bm{\lambda} \right]_j \right) < 1.
\]
The maximization over the uncertainty set and time can be performed by fixing any $t$ and maximizing over the $\hat{\bm{\tau}}(t)$ values that satisfy \eqref{eq:unc_service}. This reduces the worst-case vector load computation to solving $m$ linear optimization problems, one for each entry of $-\hat{\bm{D}}(t)\hat{\bm{A}}^{-1} \bm{\lambda}$. Ensuring that this condition holds guarantees that the vector load remains strictly below 1 for all $t \in [0, T]$ in the robust network. 

Since Problem \eqref{eq:robfluid} is also an SCLP problem, Algorithm \ref{alg:main} can be applied the same way to the robust MFNQETs. However, because Problem \eqref{eq:robfluid} includes additional variables and constraints compared to Problem \eqref{eq:fluid}, the computational burden for generating training data may be higher. See also \citep{shindin2024robust}, which develops a specialized algorithm for robust SCLPs.

In Section \ref{sec:exp_rob}, we compare the robust policies trained on robust MFQNETs with the policies trained on deterministic MFQNETs. Our results show that the robust policies reduce the need for frequent policy re-training and provide greater robustness to parameter uncertainties. However, robust policies tend to be more conservative, which leads to suboptimal performance when the level of uncertainty is low.

\section{Computational Experiments}
\label{sec:numerical experiment}

This section presents the findings of computational experiments conducted on MFQNETs with varying sizes. We analyze the accuracy of the policy learned by Algorithm \ref{alg:main} and compare its online application speed with that of the algorithm by \citet{evgenyweiss2021}. In addition, we provide insights on the optimal policy of MFQNET control problems by presenting some of the actual decision trees. We also apply OCT-H with sparsity on several MFQNET problems and analyze the impact of sparsity on the performance and the resulting decision tree. The networks in this section are taken from \citep{robustfluid} and \citep{evgenyweiss2021}. Finally, we compare the robust and the deterministic policies on the criss-cross and the Rybko-Stolyar networks described in Section \ref{sec:example}.

\subsection{Experimental Setting}
\label{sec:exp_setting}

\begin{figure}
    \centering
    \includegraphics[scale = 0.33]{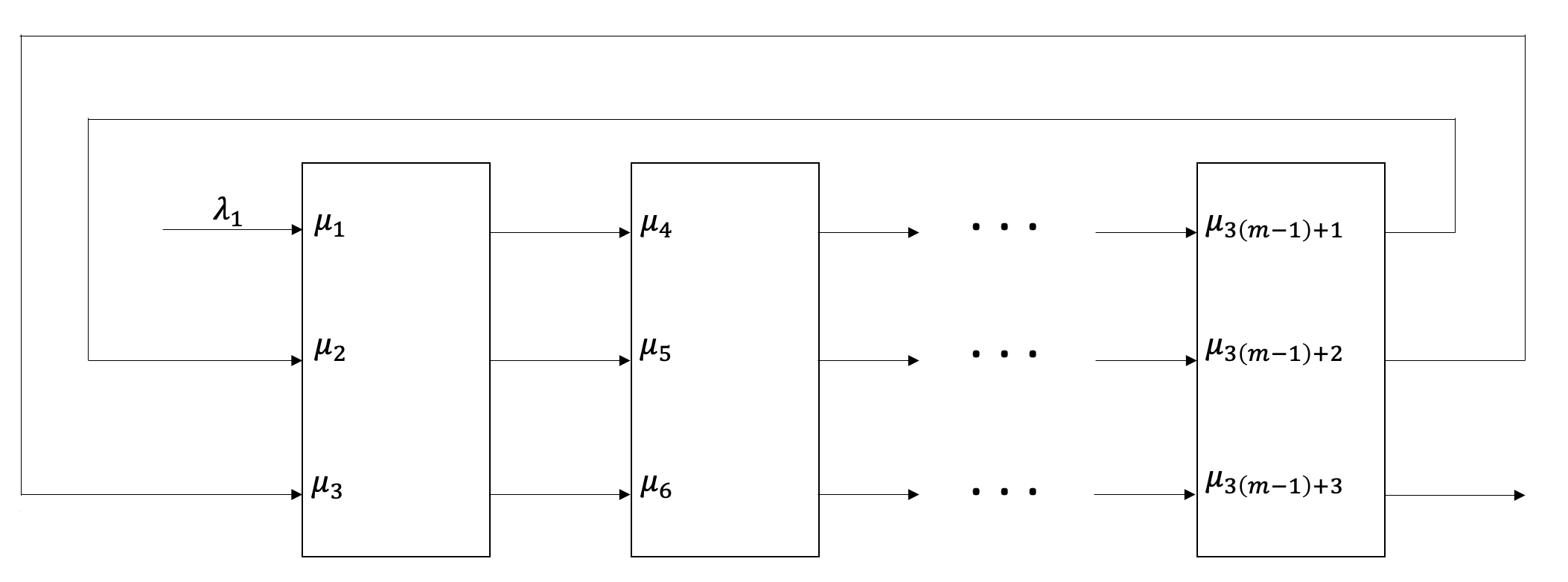}
    \caption{Reentrant network.}
    \label{fig:reentrant}
\end{figure}

We outline the experimental setting for the experiments in Sections \ref{sec:speed} to \ref{sec:sparsity}. The experimental setting for the robust policies are provided separately in Section \ref{sec:exp_rob}. For the experiments in Sections \ref{sec:speed} to \ref{sec:sparsity}, we consider a reentrant network with $m$ servers and $3m$ classes of fluids. Each Server $i \in [m]$ processes fluids of Classes $3(i-1)+1,3(i-1)+2$ and $3(i-1) + 3$. Only Class 1 fluids take external arrivals with the arrival rate $\lambda_1$. Class $3(i-1) + 1$ fluids become Class $3i + 1$ until they become Class $3(m-1) + 1$. After Class $3(m-1) + 1$ fluids are processed, they change to Class 2 and enter Server 1. Class $3(i-1)+2$ fluids become Class $3i + 2$ fluids until they become Class $3(m-1) + 2$. After  Class $3(m-1) + 2$ fluids are processed, they change to Class 3 to enter Server 1. Class $3(i-1)+3$ fluids become $3i+3$ fluids, until they become Class $3m$ and exit the system after processed. Reentrant networks have important applications in industries such as semiconductors, electronics, and automotive manufacturing, especially in microchip wafer fabrication and thin film production lines \citep{wein88, kumar_re-entrant_1993-1, 5223754}. We provide a graphical representation in Figure \ref{fig:reentrant}. The parameters $\bm{\lambda}, \bm{\mu}, \bm{c}$ are randomly generated using the software by \citet{evgenyweiss2021}.

% We also consider another set of reentrant networks with $m$ servers and $3m$ classes of fluids, but with randomly generated $\bm{D}$. In this set of networks, the sequence of stations that a fluid visits until it exits the system is different for every network we generated.

We test Algorithm \ref{alg:main} on the reentrant networks with varying $m$. {\color{black} However, as $m$ increases, training a policy that covers the entire state space $\mathbb{R}^{3m}_{\ge 0}$ becomes increasingly computationally demanding. Demonstrating such coverage at larger values of $m$ would require prohibitively extensive training and evaluation, making it impractical within the scope of our experimental study. As discussed in Remark \ref{remark:dist}, a more realistic alternative is to assume a distribution over states and train a policy tailored to that distribution, which is standard in statistical learning literature. However, results obtained under such an assumption would necessarily be tied to the specific choice of distribution.

We therefore adopt a middle-ground approach for our experiments. We select three large subsets of the state space using the partitioning scheme proposed in Section \ref{sec:algorithm}, based on whether components of the state vector are zero, and train a separate policy for each subset. Under the formalism of Section \ref{sec:algorithm}, this corresponds to defining a partition $\mathcal{P} = \big\{{s_1}, {s_2}, \dots, {s_{2^{3m}-1}}\big\}$ and learning policies for three selected cells. We always include the case in which none of the classes are empty, namely the strict interior region $\mathbb{R}^{3m}_{>0}$, and select two additional cells at random. We then evaluate the quality of the policy learned in each region separately.} 

After we fix some $s \in \mathcal{S}$, we generate a training set with  $N = 1, t_1 = 0, \mathcal{A} = \{ 0.5, 0.75, 5\}$ and $M = 10{,}000$. We generate a test set with $N = 1, t_1 = 0, \mathcal{A} = \{ 10\}$ and $M = 2000$. This ensures that the test data consists of state-control pairs that are completely distinct from the training data. We evaluate the out-of-sample classification accuracy of OCT-H by measuring the proportion of test data for which the learned policies correctly output the optimal control. For each MFQNET control instance that is used to generate test set, we also measure the time it takes to solve the instance using the algorithm by \citet{evgenyweiss2021}, and divide it by the time it takes for the trained decision tree to make a prediction.  This allows us to evaluate the speed-up that Algorithm \ref{alg:main} can offer compared to the numerical algorithm, specifically in scenarios where the only changing factor between instances is the state vector. We report the mean of the ratios rounded to the nearest integer as the relative speed-up of Algorithm \ref{alg:main}. 

Software for OCT-H is available at \citet{InterpretableAI}. We tune the maximum depth of the tree by grid searching over the list [3,5,10]. Public implementation of the algorithm by \citet{evgenyweiss2021} is available at \url{https://github.com/IBM/SCLPsolver}. When we use this implementation, we set the zero entries in $\bm{\lambda}$ to a small number $10^{-6}$ instead of $0$, as we have observed that this results in better numerical stability. The experiments were executed on a MacBook Pro with 2.6 GHz Intel Core i7 CPU and 16GB of RAM, except for the training part. We trained decision trees on MIT Engaging Computing Cluster with Dell C6300,
2 socket Intel E5-2690v4 processor, 14 Cores per CPU and 128 GB RAM.

\subsection{Speed and Accuracy}
\label{sec:speed}

\begin{table*}\centering

\begin{tabular}{cccccc}
\toprule
$m$ & $|s|$  & $|\{\bm{u}^*\}|$ & Training Time (hours:minutes) & Speed-up & Accuracy (\%)    \\ 
\midrule

\multirow{3}{*}{3}& 9 &3& 00:30 & 153 & 100 \\

& 7& 3& 00:38& 196 & 100 \\

& 5& 3& 00:48 & 125 & 100 \\

\hline

\multirow{3}{*}{7}& 21&4 & 01:26 & 255 &100  \\

& 9&4& 00:34& 151 &100 \\

& 7& 4& 00:34 &  245 & 100 \\

\hline

\multirow{3}{*}{8}& 24& 4& 00:28 & 296&100 \\

& 12 &4& 00:44& 278 & 100 \\

& 7& 4& 00:23 & 313 & 100 \\

\hline 

\multirow{3}{*}{9}& 27& 4& 00:30 &   364 &100 \\

& 24 &2& 00:33 &  360 &100  \\

& 7& 2& 00:40 &  343 & 100 \\

\hline

\multirow{3}{*}{14}& 42& 6& 03:50 & 781  &100 \\

& 36 & 6& 04:34 & 790  & 100  \\

& 7& 4& 04:20 & 661 & 100 \\

\hline

\multirow{3}{*}{20}& 60& 9& 46:20 &  700 &100 \\

& 30 &9 & 44:10 &  599 &100  \\

& 7&  9& 40:24 &  628 & 100 \\

\hline

\multirow{3}{*}{33}& 99 & 9& 48:20 & 6014 &100 \\

& 51 &9&  47:30 &  994  &100  \\

& 45& 9& 42:20 &  982  & 100 \\

\bottomrule
\end{tabular}
\caption{Experiment results for the reentrant network.}
\label{table:reentrant}
\end{table*}

In Table \ref{table:reentrant}, we report the results of numerical experiments, focusing on the speed and accuracy of Algorithm \ref{alg:main}. In the second column, we report the number of non-zero entries of the state vector, denoted by $|s|$. In the third column, we report the number of distinct labels for the classification task, denoted by $|\{\bm{u}^*\}|$. In the rest of the columns we report $m$, the training time for OCT-H, the relative speed-up of Algorithm \ref{alg:main} and the out-of-sample classification accuracy on the test set, rounded to the third decimal place.

\paragraph{Observations from Table \ref{table:reentrant}}
\begin{itemize}
    \item Algorithm 1 achieves perfect accuracy regardless of the size of the network, the number of unique labels and the number of non-zero entries. This indicates that the learned policies consistently generalize successfully to unseen states.
    \item The training time for OCT-H takes at most 48 hours in our experiment, suggesting that data generation and training might take hours to days in practice. 
    \item Once a policy is learned, Algorithm \ref{alg:main} can achieve a significant speed-up compared to the algorithm by \citet{evgenyweiss2021}, given that the only changing factor between instances is the state vector. The speed-up ranges from hundreds to thousands of times faster in our experiment. In general, this relative speed-up becomes even greater as the dimension of the state space gets higher.
    \item As the dimension of the state space gets higher, the number of distinct labels do not increase significantly. This observation suggests that even for high dimensional problems, the structure of the optimal policy might be simple enough to be learned by OCT-H with shallow decision trees.
\end{itemize}

\subsection{Interpretability}
\label{sec:interpretability}

We now provide the decision tree for a problem solved in Section \ref{sec:speed} and develop insights on the structure of the learned policy. Although not all of the node splits have straightforward interpretations, we highlight a few splits that make intuitive sense.  

Due to space concerns, we display the prediction targets in the tree figures as the list of fluid classes that are prioritized, rather than the optimal control vector $\bm{u}^*$ itself. The fluid classes that are prioritized receive effort 1, and the rest of the fluid classes receive effort 0. In addition, we assign a number to each node split and provide a separate table that contains information on the hyperplane for each split. 

Furthermore, we introduce a vector ${\bm{c}}/{\bm{\mu}}$ that offers an interesting interpretation on the learned policy. This vector captures the relative cost of holding each fluid class in terms of their service rate. A higher value in this vector indicates that the corresponding fluid class poses a greater challenge to the fluid network controller.

\begin{figure}
\centering
    \includegraphics[scale = 0.5]{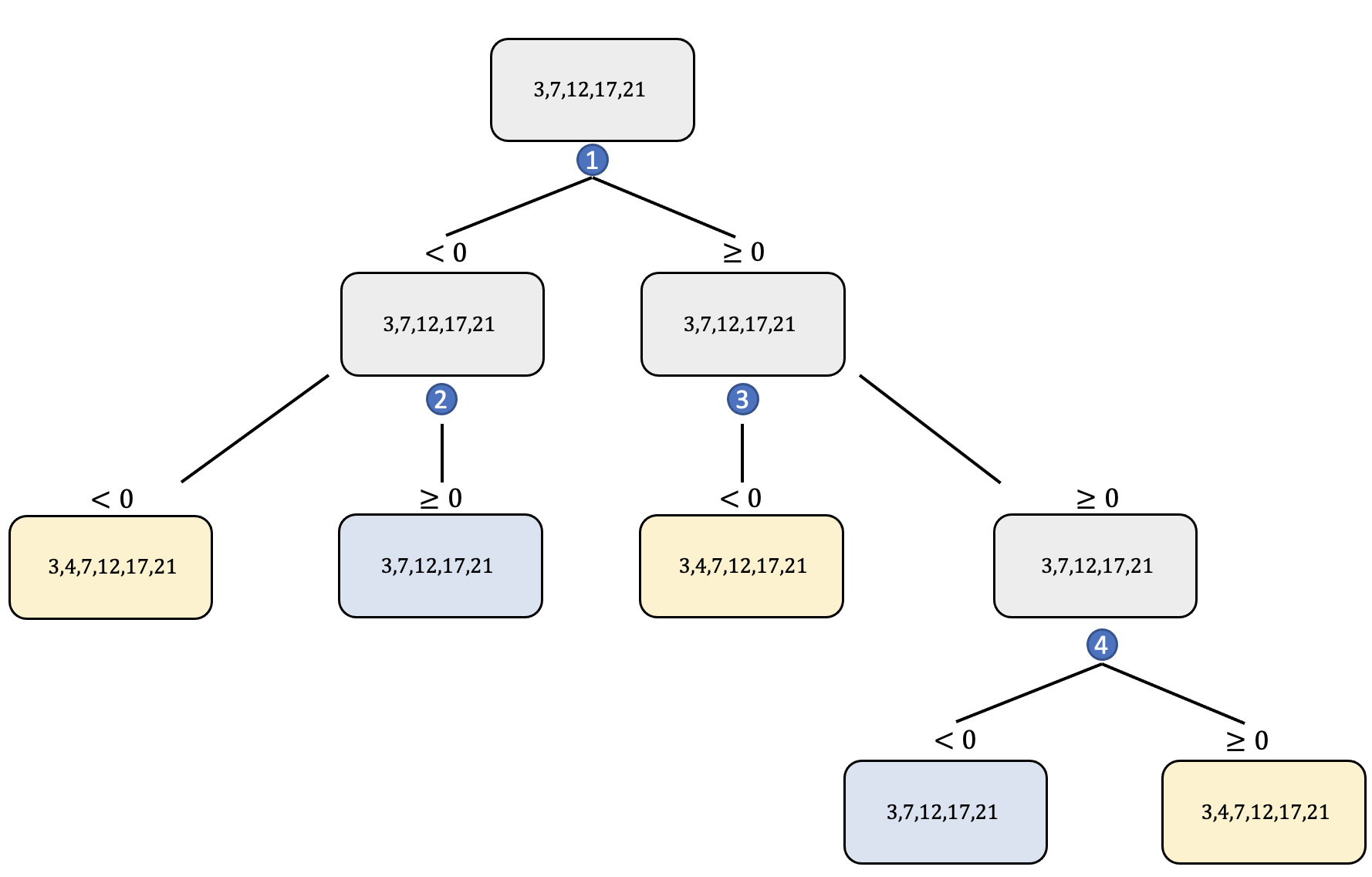}
    \caption{The decision tree learned by OCT-H for the reentrant network with $m=7$ and $|s| = 21$.}
    \label{fig:21_tree}
\end{figure}

In Figure \ref{fig:21_tree}, we provide the decision tree for the problem with $m = 7$, confined to strictly positive state vectors ($|s| = 21$). In Table \ref{table:nodesplit_reentrant_21}, we provide the node split information associated with the decision tree. For this problem, the parameters rounded to the third decimal place are
\begin{align*}
    & \bm{\mu} = (0.143, 0.253, 0.002, 0.287, 0.169, 0.278, 0.22, 0.11, 0.207, 0.216, 0.299, 0.004, 0.185, 0.205,\\
    & 0.25,0.268, 0.027, 0.028, 0.245, 0.168, 0.248), \\
    & \bm{c} = (0.705, 0.235, 0.972, 0.968, 0.719, 0.107, 1.484, 1.395, 0.493, 0.746, 1.584, 1.512, 0.07, 0.892, \\
       & 1.255, 0.305, 1.941, 1.496, 0.643, 1.021, 1.975).
\end{align*}
After we compute $\bm{c}/\bm{\mu}$ and sort it in descending order, the resulting indices in the sorted order is
$$(3,12,17,18,8,21,7,20,11,15,1,14,5,10,4,19,9,16,2,6,13).
$$

\begin{table*}\centering

\begin{tabular}{cc}
\toprule
Node split number & Hyperplane     \\ 
\midrule

1&  $0.144x_1 - 0.002x_4 + 0.17x_7 - 0.05x_{10} - 0.0005$ \\

\hline

2& $-0.096x_4 + 1.036x_7$ \\

\hline

\multirow{3}{*}{3}& $-732.7x_1 - 34.84x_{2} - 5.311x_{3} - 1989.5x_4 + 19.1x_{5}   $  \\

& $+ 14304.6x_7 + 4.175x_{8} + 2939.7x_{10} +33.72x_{11}+ 62.25x_{12}$ \\

& $- 6.67x_{13} - 12.42x_{14} + 1.382x_{15} + 8.84x_{16} - 2.794x_{19} + 12.92x_{20}$ \\

\hline

4& $-2711.2x_3 + 0.5262x_{12} - 0.0001$  \\

\bottomrule
\end{tabular}
\caption{Node split information on the decision tree in Figure \ref{fig:21_tree}.}
\label{table:nodesplit_reentrant_21}
\end{table*}

\paragraph{Observations from Figure \ref{fig:21_tree}}

\begin{itemize}
    \item Class 3,7,12,17,21 fluids are always prioritized, regardless of the node. These fluid classes are often the highest ranking classes within their respective server in terms of the value in the vector ${\bm{c}}/{\bm{\mu}}$. The only exception is Class 8, as Class 8 is not prioritized even though it is ranked the highest in its server. This observation implies that the learned policy for this network is to drain the ``toughest"  fluid classes from the system first.
    \item The only difference in the nodes is whether to process Class 4 fluids or idle Server 2. See split 1 and 2, for example. If $x_4$ is relatively large compared to a linear combination of $x_1, x_7, x_{10}$, and if $x_4$ is again relatively large compared to $x_7$, the decision is to process Class 4 fluids instead of idling Server 2. However, after traversing the left edge in split 1, if $x_7$ turns out to be too large compared to $x_4$, Class 4 fluids are not processed. A possible explanation is that as Class 4 fluids become Class 7 after processed, it might be beneficial to idle Server 2 in case there are too many Class 7 fluids waiting in the queue.
    \item See split 3. If we focus on the terms associated with $x_4$ and $x_7$, again the decision is to process Class 4 fluids if $x_4$ is relatively large compared to $x_7$. The same interpretation as above can be applied to this decision.
    
\end{itemize}

\subsection{OCT-H with sparsity}
\label{sec:sparsity}

In this experiment, we apply OCT-H with sparsity instead of OCT-H in Algorithm \ref{alg:main} on a subset of the problems solved in Section \ref{sec:speed}. As mentioned in Section \ref{sec:oct}, OCT-H with sparsity often results in more interpretable decision trees compared to OCT-H. The purpose of this experiment is to analyze the price we have to pay in order to gain more interpretability. We vary the proportion of the total number of states allowed to be used for splits, denoted by sparsity parameter. We analyze how the sparsity parameter affects the training time and the classification accuracy on the test set. Table \ref{table:reentrant_sparsity} provides the experiment results, where the same notations as Table \ref{table:reentrant} are used. We summarize our findings in the following.

\begin{figure}
\centering
    \includegraphics[scale = 0.5]{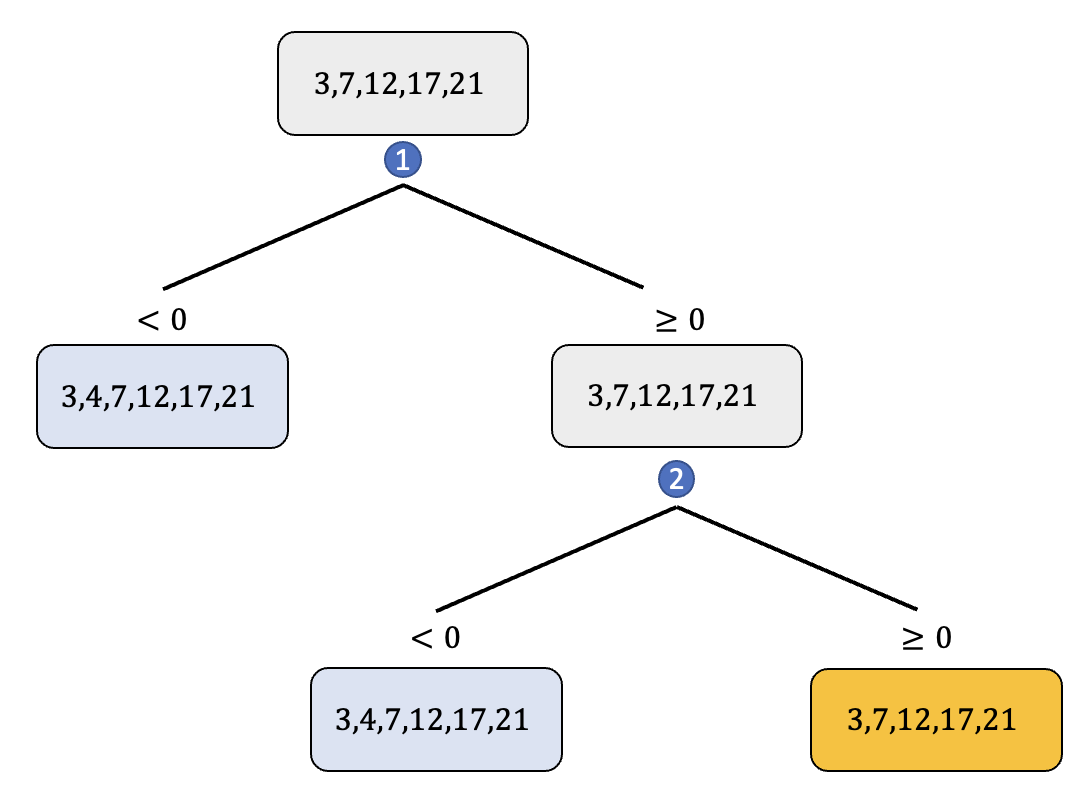}
    \caption{The decision tree learned by OCT-H with the sparsity parameter 0.25 for the reentrant network with $m=7$ and $|s| = 21$.}
    \label{fig:21cl_sparse}
\end{figure}

\begin{table*}\centering

\begin{tabular}{cc}
\toprule
Node split number & Hyperplane     \\ 
\midrule

1& $-0.316x_4 + 3.398x_7$  \\

\hline

2& $-0.034x_1 - 0.103x_4 + 0.753x_7 + 0.152x_{10} + 0.0003x_{11}$ \\

\bottomrule
\end{tabular}
\caption{Node split information on the decision tree in Figure \ref{fig:21cl_sparse}.}
\label{table:nodesplit_reentrant_21_sparse}
\end{table*}

\begin{table*}\centering

\begin{tabular}{ccccc}
\toprule
$m$ & $|s|$ & Sparsity Parameter  & Training Time (hours:minutes) & Accuracy (\%)    \\ 
\midrule

\multirow{2}{*}{7}& \multirow{2}{*}{21}& 0.5 & 00:48 & 99.8   \\

&  & 0.25 &  00:35 & 99.8  \\

\hline 

\multirow{2}{*}{9}& \multirow{2}{*}{27} & 0.5& 00:08 &   100  \\

&  & 0.25 & 00:07 & 100  \\

\hline

\multirow{2}{*}{14}& \multirow{2}{*}{42}& 0.5 & 02:52 & 98.3   \\

&  & 0.25 & 00:58 &   97 \\

\hline

\multirow{2}{*}{20}& \multirow{2}{*}{60}& 0.5 & 27:28 &  95.3  \\

&  & 0.25 & 14:20 & 94   \\

\hline

\multirow{2}{*}{33}& \multirow{2}{*}{99} &  0.5 & 26:51 &  94 \\

&  & 0.25 &  14:14 &  94    \\

\bottomrule
\end{tabular}
\caption{Experiment results for the reentrant network using OCT-H with sparsity.}
\label{table:reentrant_sparsity}
\end{table*}

\paragraph{Observations from Table \ref{table:reentrant_sparsity} }

\begin{itemize}
    \item In general, classification accuracy slightly degrades as sparsity parameter gets smaller. However, classification accuracy never gets below 94\% in our experiment, suggesting that OCT-H with sparsity can still learn high-quality policies.
    \item Training becomes faster as the sparsity parameter gets smaller. For the sparsity parameter 0.25, training can be around 4 times faster than OCT-H.
\end{itemize}

We provide the decision tree for the problem with $m=7, s = 21$ and the sparsity parameter 0.25 in Figure \ref{fig:21cl_sparse}. We compare this tree with the tree in Figure \ref{fig:21_tree}, which is learned by OCT-H on the same problem. Note that OCT-H with sparsity achieves 99.8 \% accuracy on this problem, which is only 0.2 \% decrease compared to OCT-H. Node split information is given in Table \ref{table:nodesplit_reentrant_21_sparse}.  

\paragraph{Observations from Figure \ref{fig:21cl_sparse} }

\begin{itemize}
    \item The states used for the splits are a strict subset of the states used for the splits in Figure \ref{fig:21_tree}. 
    \item The learned policy is also qualitatively similar to the policy learned with OCT-H. For example, in split 1, if $x_4$ is relatively large compared to $x_7$, the decision is to process Class 4 fluids. In split 2, if we focus on the terms associated with $x_4$ and $x_7$, again the decision is to process Class 4 fluids if $x_4$ is relatively large compared to $x_7$. Else, we idle Server 2 so that the queue on Class 7 fluids do not increase.
\end{itemize}

\subsection{Comparison of Robust and Deterministic Policies}
\label{sec:exp_rob}

We train deterministic policies on the Rybko-Stolyar (Figure \ref{fig:rybko}) and the criss-cross networks (Figure \ref{fig:crisscross}), using the same parameters as described in Section \ref{sec:example}. We train robust policies by solving the robust MFQNET control problems (Problem \eqref{eq:robfluid}) on these networks. In the robust problems, we set $\Gamma = 0.1$ for the criss-cross network and $\Gamma = 0.25$ for the Rybko-Stolyar network. For both networks, we set $\tilde{\bm{\tau}} = 0.25\bar{\bm{\tau}}$, where $\bar{\bm{\tau}}$ denotes the nominal parameter used in the deterministic models. 

Previously, in Section \ref{sec:example}, we focused on learning an optimal policy for the criss-cross network only within the interior of the state space. For this experiment, however, we train a single decision tree that covers the entire state space, including the boundaries. Other training details follow the same procedure described in Section \ref{sec:exp_setting}.

After the training is complete, we apply these policies to test instances with varying service rates (or equivalently, varying service times) and initial states to compute the resulting objective costs. Each test instance is a deterministic MFQNET control problem, where the service time is generated uniformly at random from the interval $[\bar{\bm{\tau}}, (1+\epsilon)\bar{\bm{\tau}}]$. Here, $\epsilon$ represents the noise level, and we vary its value to compare the deterministic and robust policies under different noise conditions. We use different values of $\epsilon$ for the criss-cross and the Rybko-Stolyar networks, because these two networks exhibit different levels of sensitivity to noise. For each value of $\epsilon$, we generate 100 test instances. The initial state for each test instance is sampled uniformly at random from the sphere with radius 10. We set the time horizon $T$ to be the draining time under the optimal trajectory. 

For each test instance, we compare the optimal objective cost with the costs achieved by applying the deterministic and the robust policies. To simulate the continuous dynamics, we discretize the system using a step size of 0.0001 and compute the integral objective costs. To determine the suboptimality of a policy, we subtract the optimal objective cost from the cost achieved by the policy and divide this difference by the optimal objective cost. We then report both the maximum and mean suboptimality values across all test instances generated with the same $\epsilon$.

As in the experiment in Section \ref{sec:speed}, we also report the out-of-sample classification accuracies of the learned policies to empirically verify Corollaries \ref{th:oct} and \ref{th:rob_oct}. Each test set consists of 2000 unseen state-control pairs. Note that these test sets are not generated from networks with perturbed service rates. The test sets for the deterministic policies are generated from the deterministic MFQNETs with nominal service rates, while the test sets for the robust policies are generated from the same robust MFQNETs that are used to train these policies.

When applying either the deterministic or the robust policies to networks with perturbed service rates, caution is required to avoid infeasibility. First, if $x_i(t) = 0$ for some class $i \in [n]$, additional constraints on the control variable should be imposed to ensure that $x_i(t)$ does not become negative. For deterministic MFQNETs, these constraints, derived from the state equations, are defined as:
$$
\dot{x}_i(t) \equiv [\bm{Au}(t) + \bm{\lambda}]_i \geq 0, \quad \text{if } x_i(t) = 0, \forall i \in [n].
$$ 
These constraints depend on the problem parameters $\bm{\lambda}$ and $\bm{\mu}$. Hence, an optimal policy for one network may result in an infeasible state trajectory for a network with slightly different parameters. To avoid this, when $x_i(t) = 0$, we project the control outputs of policies into the polyhedron defined by these constraints. This projection, which finds the closest element in the polyhedron with respect to the $\ell^2$ distance, is typically not computationally expensive. It involves solving a convex quadratic optimization problem with at most $n$ linear constraints. A similar projection of $\hat{u}_i(t)$ can be defined for the robust policy.

Second, recall that when applying a robust policy, we must divide the control output $\hat{\bm{u}}$ by $\bm{\mu}$ to make it equivalent to $\bm{u}$. Depending on the value of $\bm{\mu}$, the sum of the resulting control values for a single server might exceed 1. If this occurs, we further normalize these control values by their sum to ensure that the scaled total never exceeds 1.

In Table \ref{table:rob_rs} and \ref{table:rob_cc}, we report the results of numerical experiments.

\begin{table*}\centering

\begin{tabular}{ccccc}
\toprule
Policy &  Accuracy (\%) & Noise Level ($\epsilon$)   & Max Suboptimality & Mean Suboptimality \\ 
\midrule

\multirow{3}{*}{Deterministic} & \multirow{3}{*}{100} & 0 & 0.0047 &  0.0009\\

&  & 0.05 & 0.1458 & 0.0354 \\

 &  & 0.10 & 0.3483  & 0.0811  \\

\hline

\multirow{3}{*}{Robust} & \multirow{3}{*}{100} & 0 & 0.0425 & 0.0024 \\

 &  & 0.05 & 0.0884 & 0.0281\\

 &  & 0.10  & 0.1750& 0.0408  \\

\bottomrule
\end{tabular}
\caption{Comparison of robust and deterministic policies on the Rybko-Stolyar network.}
\label{table:rob_rs}
\end{table*}

\begin{table*}\centering

\begin{tabular}{ccccc}
\toprule
Policy & Accuracy (\%) & Noise Level ($\epsilon$)  &  Max Suboptimality & Mean Suboptimality \\ 
\midrule

\multirow{3}{*}{Deterministic} & \multirow{3}{*}{100} & 0 & 0.0023 &  0.0009\\

&  & 1 & 0.0229 & 0.0051 \\

 &  & 2 & 0.8011 & 0.1260  \\

\hline

\multirow{3}{*}{Robust} & \multirow{3}{*}{100} & 0 & 0.1085 & 0.0991 \\

 &  & 1 & 0.0362 & 0.0058\\

 &  & 2  & 0.3392 & 0.0705  \\

\bottomrule
\end{tabular}
\caption{Comparison of robust and deterministic policies on the criss-cross network.}
\label{table:rob_cc}
\end{table*}

\paragraph{Observations from Tables \ref{table:rob_rs} and \ref{table:rob_cc}}
\begin{itemize}
    \item The out-of-sample classification accuracies for the robust policies consistently reach 100\%. This demonstrates that Algorithm \ref{alg:main} effectively learns an optimal policy for robust MFQNETs.
    \item  The suboptimalities of the deterministic policies are very small on the nominal problems generated with $\epsilon = 0$. On average, suboptimalities are below 0.001, with a maximum of 0.0047 (note that achieving zero suboptimality is challenging due to approximations inherent in discretizing the continuous dynamics).  Along with the accuracy results in Section \ref{sec:speed}, this demonstrates that the deterministic policies learned using Algorithm  \ref{alg:main} are near-optimal. 
    \item The suboptimalities of deterministic policies remain relatively low when the noise level is low. For example, in the criss-cross network, the maximum suboptimality is at most 2.3\% even when the parameter deviation equals the nominal service time ($\epsilon = 1$). However, for both networks, as the noise level increases, the suboptimalities of deterministic policies grow larger. 
    \item For the nominal problems, the deterministic policies outperforms the robust policies. However, as the noise level increases, robust policies begin to outperform the deterministic policies. This indicates that the robust policies are more conservative, leading to suboptimal performance when the level of uncertainty is low. At the same time, they are more resilient to parameter perturbations, which is a clear advantage in more uncertain environments.
\end{itemize}

\section{Conclusions}
\label{sec:conclusion}

We presented an approach to solve MFQNET control problems using OCT-H. We proved that MFQNET control problems have piecewise constant optimal policy, where the segments are separated by hyperplanes passing through the origin. Based on this result, we developed an algorithm to use OCT-H to learn the optimal policy of MFQNET control problems. Computational experiments demonstrate that OCT-H can learn empirically optimal policies of MFQNET control problems with varying sizes.  Once the learning is complete, Algorithm \ref{alg:main} can achieve significant speed-up compared to the algorithm by \citet{evgenyweiss2021}, in scenarios where the state vector is the only factor that differs between instances. Furthermore, we demonstrated that the simple decision tree structure enables us to develop insights on large dimensional MFQNET control problems.

% Appendix here
% Options are (1) APPENDIX (with or without general title) or
%             (2) APPENDICES (if it has more than one unrelated sections)
% Outcomment the appropriate case if necessary
%

\appendix
\section{Additional Experiments}
{\color{black}
We study the effect of the training set size for Algorithm~\ref{alg:main} in order to understand how many training instances are required to achieve the perfect classification accuracy observed in our main experiments. For this study, we fix $m = 7$ and focus on the reentrant network setting considered in Section~\ref{sec:speed}. In Section~\ref{sec:speed}, we use a total of 40{,}000 training instances for all settings, obtained by augmenting an initial set of 10{,}000 training samples three times. Here, we vary the number of training instances to assess the sensitivity of the learned policy to the size of the training dataset.

We also consider a more challenging variant of the experiment in which we modify the partition used in Section~\ref{sec:speed} for $m = 7$ by combining the three cells into a single region. We train a single decision tree on the combined data, requiring the tree to cover a substantially larger state space than in the partition used in the main experiment. For this combined-cell experiment, the total number of training instances is distributed evenly across the original cells. Test data are generated analogously to the procedure described in Section~\ref{sec:exp_setting}, by sampling test instances from each cell and then aggregating them.

The results are reported in Table~\ref{table:training_ablation}. With a slight abuse of notation, we use $21 \cup 9 \cup 7$ to denote the setting in which training data corresponding to $|s| = 21$, $9$, and $7$ are combined and a single decision tree is trained on the aggregated dataset. Overall, we observe that classification accuracy remains extremely high across all settings, though it may fall slightly below 100\% when the training set is very small. Importantly, even in the combined-partition setting, perfect classification accuracy is achieved once the training dataset is sufficiently large.

\begin{table*}\centering

\begin{tabular}{ccccc}
\toprule
 $|s|$   & Training Set Size &  Training Time (hours:minutes) & Accuracy (\%)    \\ 
\midrule

 \multirow{4}{*}{21} & 2,000 & 00:02 & 98 \\

   & 4,000 & 00:05 & 99 \\

    & 8,000 & 00:17 & 100 \\

    & 20,000 & 00:52  & 100 \\

    \hline

\multirow{4}{*}{9} & 2,000 & 00:02  & 94 \\

   & 4,000 & 00:06 & 97 \\

    & 8,000 & 00:15 & 97 \\

     & 20,000 & 00:30 & 99 \\

      \hline

 \multirow{4}{*}{7} & 2,000 & 00:01 & 99 \\

   & 4,000 & 00:02 & 100 \\

    & 8,000 & 00:04 & 100 \\

     & 20,000 & 00:21 & 100 \\

      \hline

       $21 \cup 9 \cup 7 $  & 24,000 & 01:23 & 98 \\

 $21 \cup 9 \cup 7 $  & 60,000 &  05:45 & 100 \\

  $21 \cup 9 \cup 7 $  & 120,000 & 17:22 & 100 \\

\bottomrule
\end{tabular}
\caption{Effect of training set size on classification accuracy for the reentrant network ($m=7$).}
\label{table:training_ablation}
\end{table*}

Next, we report the standalone runtime of the numerical solution algorithm of \citet{evgenyweiss2021}, which is used to generate training data across all experimental settings considered in Section~\ref{sec:numerical experiment}. Tables~\ref{table:runtime} and~\ref{table:robust_time} summarize the average runtime required to solve a single instance of Problem~\eqref{eq:fluid}, as well as its robust variants, for each configuration reported in Tables~\ref{table:reentrant}, \ref{table:rob_rs}, and~\ref{table:rob_cc}.

\begin{table*}\centering
\begin{minipage}[t]{0.49\textwidth}\centering
\vspace{0pt}
\begin{tabular}{ccc}
\toprule
$m$ & $|s|$ & Avg Solver Time (seconds) \\
\midrule
\multirow{3}{*}{3}  & 9  &  0.0038 \\
                   & 7  &  0.0060 \\
                   & 5  &  0.0039 \\
\midrule
\multirow{3}{*}{7}  & 21 & 0.0081 \\
                   & 9  &  0.0056 \\
                   & 7  &  0.0085 \\
\midrule
\multirow{3}{*}{8}  & 24 & 0.0137 \\
                   & 12 & 0.0091 \\
                   & 7  &  0.0103 \\
\midrule
\multirow{3}{*}{9}  & 27 & 0.0158  \\
                   & 24 & 0.0139 \\
                   & 7  &  0.0142 \\
\bottomrule
\end{tabular}
\end{minipage}
\hspace{-4em}
\begin{minipage}[t]{0.49\textwidth}\centering
\vspace{0pt}
\begin{tabular}{ccc}
\toprule
$m$ & $|s|$ & Avg Solver Time (seconds) \\
\midrule
\multirow{3}{*}{14} & 42 & 0.0332 \\
                   & 36 &  0.0268 \\
                   & 7  & 0.0206 \\
\midrule
\multirow{3}{*}{20} & 60 & 0.0699 \\
                   & 30 & 0.0459 \\
                   & 7  &  0.0486 \\
\midrule
\multirow{3}{*}{33} & 99 & 1.0237  \\
                   & 51 & 0.6745 \\
                   & 45 & 0.6105 \\
\bottomrule
\end{tabular}
\end{minipage}
\caption{Average per-instance solver runtime for the reentrant network.}\label{table:runtime}
\end{table*}

\begin{table*}\centering
\begin{minipage}[t]{0.49\textwidth}\centering
\vspace{0pt}
\begin{tabular}{cc}
\toprule
Network Type & Avg Solver Time (seconds) \\
\midrule
Criss-cross   & 0.0036  \\
\bottomrule
\end{tabular}
\end{minipage}\hfill
\begin{minipage}[t]{0.49\textwidth}\centering
\vspace{0pt}
\begin{tabular}{cc}
\toprule
Network Type & Avg Solver Time (seconds) \\
\midrule
Rybko-Stolyar & 0.0022  \\

\bottomrule
\end{tabular}
\end{minipage}
\caption{Average per-instance solver runtime for the robust MFQNETs.}
\label{table:robust_time}
\end{table*}
}

%
%   or
%
% \appendix
% \section{<Title of Section A>}
% \section{<Title of Section B>}
% etc
% 

%%
%\theendnotes

% Acknowledgments here
%\ACKNOWLEDGMENT{}

% References here (outcomment the appropriate case)

% CASE 1: BiBTeX used to constantly update the references
%   (while the paper is being written).
\FloatBarrier
\bibliographystyle{abbrvnat}
\bibliography{reference}

% CASE 2: BiBTeX used to generate mypaper.bbl (to be further fine tuned)
%\input{mypaper.bbl} % outcomment this line in Case 2

%If you don't use BiBTex, you can manually itemize references as shown below.

%%%%%%%%%%%%%%%%%
\end{document}